\documentclass[lettersize,journal]{IEEEtran}

\usepackage{booktabs} %

\ifCLASSOPTIONcompsoc
  \usepackage[nocompress]{cite}
\else
  \usepackage{cite}
\fi

\usepackage[ruled]{algorithm2e} %
\usepackage{graphicx}
\usepackage{subcaption}
\usepackage{makecell}
\usepackage{colortbl}  
\usepackage{array} 
\usepackage {amsmath}

\SetAlFnt{\small}
\SetAlCapFnt{\small}
\SetAlCapNameFnt{\small}
\SetAlCapHSkip{0pt}
\usepackage{amsmath,amsfonts}
\usepackage{algorithmic}
\usepackage{array}
\usepackage{textcomp}
\usepackage{stfloats}
\usepackage{url}
\usepackage{verbatim}
\usepackage{graphicx}
\usepackage{cite}
\usepackage[hidelinks]{hyperref}
\usepackage{tabularx}
\usepackage{hyperref}

\hypersetup{
    colorlinks=true,   
    citecolor=blue
}

\newcommand{\zdx}[1]{{\color{black}{#1}}}
\newcommand{\zdxn}[1]{{\color{black}{#1}}}
\newcommand{\yyj}[1]{{\color{black}{#1}}}

\newcommand{\lfl}[1]{{\color{black}{#1}}}

\newcommand{\name}{StylizedGS\xspace}

\begin{document}
\title{\name: Controllable Stylization for \\ 3D Gaussian Splatting}

\author{Dingxi~Zhang, %
        Yu-Jie~Yuan,
        Zhuoxun~Chen, 
        Fang-Lue~Zhang, 
        Zhenliang~He,
        Shiguang~Shan,
        and~Lin~Gao
\IEEEcompsocitemizethanks{
\IEEEcompsocthanksitem This work was supported by National Natural Science Foundation of China (No.62322210), Beijing Municipal Science and Technology Commission (No.Z231100005923031), Innovation Funding of ICT, CAS (No.E461020), and VUW Faculty Strategic Research Grant (No. FSRG-ENGRADI-12684). \textit{(Corresponding author: Lin Gao.)}
\IEEEcompsocthanksitem Dingxi Zhang and Zhuoxun Chen are with the University of Chinese Academy of Sciences, Beijing, China (email: zhangdingxi20a@mails.ucas.ac.cn, chenzhuoxun20@mails.ucas.ac.cn).
\IEEEcompsocthanksitem Yu-Jie Yuan and Lin Gao are with the Beijing Key Laboratory of Mobile Computing and Pervasive Device, Institute of Computing Technology, Chinese Academy of Sciences, Beijing, China, and also with the University of Chinese Academy of Sciences, Beijing, China (email: yuanyujie66@gmail.com, gaolin@ict.ac.cn).
\IEEEcompsocthanksitem Fang-Lue Zhang is with Victoria University of Wellington, New Zealand (email: fanglue.zhang@vuw.ac.nz).
\IEEEcompsocthanksitem Zhenliang He and Shiguang Shan are with the Key Laboratory of Intelligent Information Processing, Institute of Computing Technology, Chinese Academy of Sciences, Beijing, China, and also with the University of Chinese Academy of Sciences, Beijing, China (email: hezhenliang@ict.ac.cn, sgshan@ict.ac.cn).
\IEEEcompsocthanksitem Our project page: \href{https://kristen-z.github.io/stylizedgs}{https://kristen-z.github.io/stylizedgs}.
}
}

\markboth{IEEE TRANSACTIONS ON PATTERN ANALYSIS AND MACHINE INTELLIGENCE, VOL x, No. x, August 2025}%
{How to Use the IEEEtran \LaTeX \ Templates}

\maketitle

\begin{abstract}
As XR technology continues to advance rapidly, 3D generation and editing are increasingly crucial. Among these, stylization plays a key role in enhancing the appearance of 3D models. By utilizing stylization, users can achieve consistent artistic effects in 3D editing using a single reference style image, making it a user-friendly editing method. However, recent NeRF-based 3D stylization methods encounter efficiency issues that impact the user experience, and their implicit nature limits their ability to accurately transfer geometric pattern styles.
Additionally, the ability for artists to apply flexible control over stylized scenes is considered highly desirable to foster an environment conducive to creative exploration. To address the above issues, we introduce \name, an efficient 3D neural style transfer framework with adaptable control over perceptual factors based on 3D Gaussian Splatting representation. We propose a filter-based refinement to eliminate floaters that affect the stylization effects in the scene reconstruction process. The nearest neighbor-based style loss is introduced to achieve stylization by fine-tuning the geometry and color parameters of 3DGS, while a depth preservation loss with other regularizations is proposed to prevent the tampering of geometry content. Moreover, facilitated by specially designed losses, \name enables users to control color, stylized scale, and regions during the stylization to possess customization capabilities.
Our method \lfl{achieves} high-quality stylization results characterized by faithful brushstrokes and geometric consistency with flexible controls. 
Extensive experiments across various scenes and styles demonstrate the effectiveness and efficiency of our method concerning both stylization quality and inference speed. 
\end{abstract}


\begin{figure*}
    \centering
    \includegraphics[width=\linewidth]{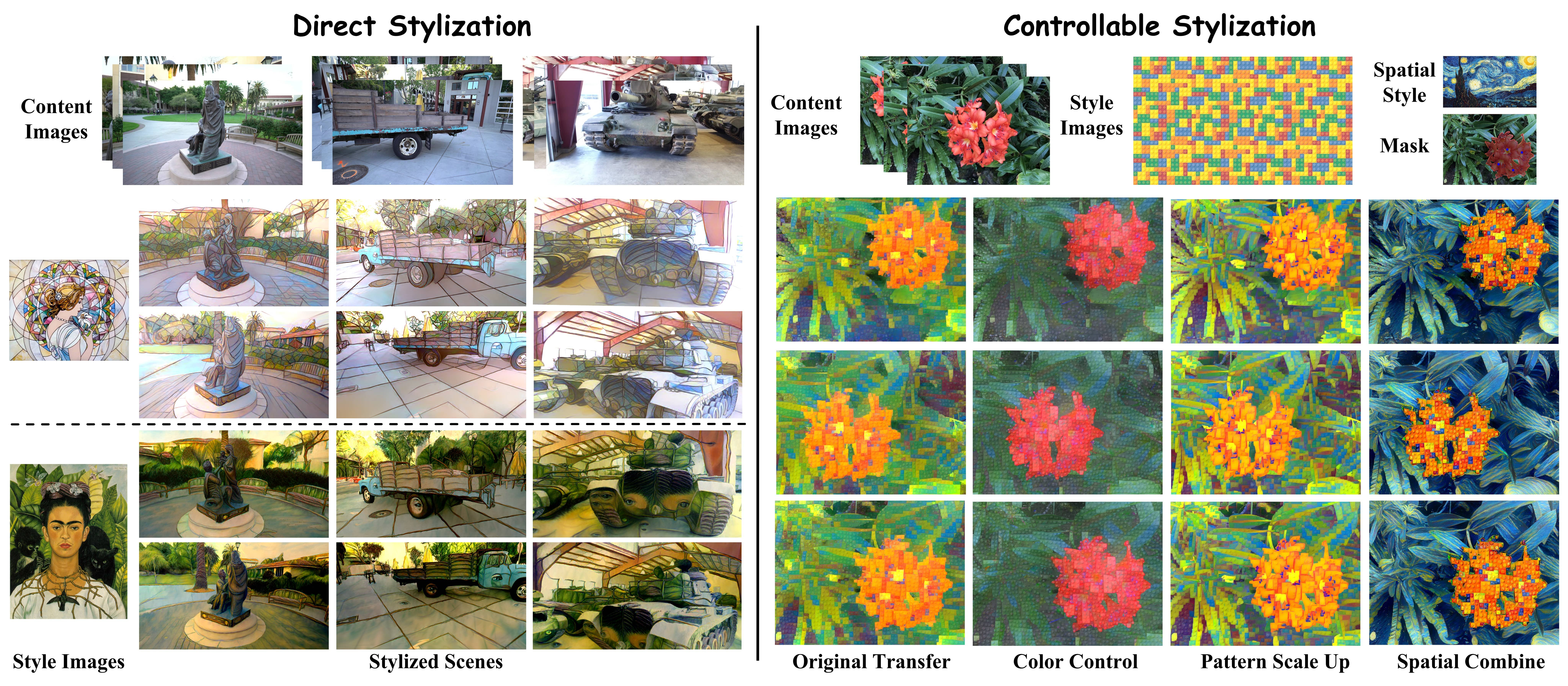}
    \caption{\textbf{Stylization Results.} Given a 2D style image, the proposed \name method can stylize the pre-trained 3D Gaussian Splatting to match the desired style with detailed geometric features and satisfactory visual quality within a few minutes. We also enable users to control several perceptual factors, such as color, the style pattern size (scale), and the stylized regions (spatial), during the stylization to enhance the customization capabilities. }
    \label{fig:teaser}
\end{figure*}

\section{Introduction}

\IEEEPARstart{N}{owadays}, the once professionally-dominated domain of artistic content creation has become increasingly accessible to novice users, thanks to recent groundbreaking advancements in visual artistic stylization research. As a pivotal artistic content generation tool in crafting visually engaging and memorable experiences, 3D scene stylization has attracted growing research efforts. 
Previous methods \lfl{have achieved attractive 3D scene style transfer results over diverse} explicit representations such as mesh~\cite{Kato_2018_CVPR, Michel_2022_CVPR, yin20213dstylenet}, voxel~\cite{guo2021volumetric, klehm2014property}, and point cloud~\cite{cao2020psnet, huang2021learning, bae2023point}. 
However, the quality of their results is limited by the \lfl{precision} of the geometric reconstructions. 
The recent 3D stylization methods benefit from the emerging implicit neural representations~\cite{chiang2022stylizing, nguyen2022snerf, fan2022unified}, such as \lfl{Neural Radiance Fields} (NeRF)~\cite{huang2022stylizednerf, zhang2022arf, wang2023nerf, pang2023locally, zhang2023transforming}, achieving more faithful and consistent stylization within 3D scenes. Nonetheless, NeRF-based methods are computationally intensive to optimize and suffer from the geometry artifacts \lfl{of} the original radiance fields.

The recently introduced 3D Gaussian Splatting (3DGS)~\cite{kerbl3Dgaussians}, showcasing remarkable 3D reconstruction quality from multi-view images with high efficiency, suggests representing the 3D scene using an array of colored and explicit 3D Gaussians. 
\lfl{Performing 3D stylization on scene represented by 3DGS} brings the benefits of prompt response and flexible style control to 3D stylization applications \yyj{due to the explicit nature of 3DGS}. 
Recent 3DGS scene manipulation methods~\cite{gaussian_grouping, GaussianEditor, chen2023gaussianeditor, tang2023dreamgaussian} explore the editing and control of 3D Gaussians using text prompts within designated regions of interest or semantic tracing. However, these approaches are driven by text input and fall short of delivering detailed style transfer capabilities. \yyj{Some styles are also difficult to describe through simple text prompt.}
Recently appeared studies on 3DGS stylization ~\cite{liu2024stylegaussian, saroha2024gaussian} show limited capabilities of learning and transferring style features. Notably, these works also lack comprehensive control over stylization effects.

In this paper, we introduce the first \textit{controllable} scene stylization method based on 3DGS, \name.  
\lfl{Given a} reference style image, our method effectively transfer \lfl{the} style features to the entire 3D scene represented by a set of 3D Gaussians. \yyj{Our method} facilitates the artistic creation of visually coherent novel views that exhibit transformed and detailed style features in a visually reasonable manner. 
More importantly, \name operates at a \lfl{fast} inference speed, ensuring efficiency when \lfl{rendering} stylized scenes. 
\lfl{To achieve effective stylization and avoid geometry distortion, a straightforward approach is to fine-tune the color of 3D Gaussians, minimizing both style loss and content loss.
However, focusing solely on color fails to effectively capture the overall stylistic features when the style patterns contain intricate details.
}
Instead, we propose formulating the 3DGS stylization as a joint optimization of both the geometry and color of 3D Gaussians to capture the detailed style features while preserving the semantic content. Specifically, our method \lfl{employs} a two-stage stylization framework. In the first stage, we optimize the 3D Gaussians to align the color distribution with the style image \lfl{while reducing} artifacts and geometry noises \lfl{using} our filter-based 3D Gaussian refinement. \lfl{In the second stage, we achieve} final stylization \lfl{by applying} a nearest neighbor feature match (NNFM) loss, \lfl{optimizing the 3D Gaussians' parameters to capture} coherent fine-grained style details. 
To \lfl{enable} users to customize the stylization process, we propose a series of strategies and loss functions to control perceptual factors such as color, scale, and spatial regions ~\cite{gatys2017controlling, jing2018stroke, castillo2017zorn, li2023arf} in the final stylized results. 
\lfl{By incorporating this enhanced} level of perceptual controllability, our approach \lfl{synthesizes diverse 3D scene stylization results with varied visual characteristics.}
Furthermore, we introduce a depth preservation loss that eliminates the necessity for additional networks or regularization operations to \lfl{preserve} the learned 3D scene geometry. It establishes a delicate balance between geometric optimization and depth preservation, facilitating effective style pattern transfer while mitigating significant deterioration of geometric content.

Our contribution can be summarized as follows:
\begin{itemize}
    \item  We introduce \name, a novel controllable 3D Gaussian stylization method that organically integrates the filter-based 3DGS refinement and the depth preservation loss with other stylization losses to transfer detailed style features and produce faithful novel stylized views. 
    \item We empower users with an efficient stylization process and flexible control through specially designed learning schemes and losses, enhancing their creative capabilities.
    \item Our approach achieves significantly reduced training and rendering times while generating high-quality stylized scenes compared with existing 3D stylization methods.
\end{itemize}

\vspace{-3mm}

\section{Related Work}
\textbf{Image Style Transfer.} 
Style transfer aims to generate synthetic images with the artistic style of given images while preserving content. Initially proposed in neural style transfer methods~\cite{gatys2015neural, gatys2016image}, this process involves iteratively optimizing the output image using Gram matrix loss and content loss calculated from VGG-Net~\cite{simonyan2014very} extracted features.
Subsequent works~\cite{risser2017stable, gu2018arbitrary, kolkin2019style, liao2017visual} have explored alternative style loss formulations to enhance semantic consistency and capture high-frequency style details such as brushstrokes.
Feed-forward transfer methods~\cite{an2021artflow, huang2017arbitrary, park2019arbitrary}, where neural networks are trained to capture style information from the style image and transfer it to the input image in a single forward pass, ensuring fast stylization. Recent improvements in style loss~\cite{kolkin2019style, liao2017visual, zhang2022arf} involve replacing the global Gram matrix with the nearest neighbor feature matrix, improving texture preservation.
Some methods adopt patch matching for image generation, like PatchMatch~\cite{barnes2009patchmatch} and Fast PatchMatch~\cite{chen2016fast}, \yyj{but can only be applied to limited views.} 
Recent works~\cite{Zhang_2023_inst, he2024multi} explore popular tools such as CycleGAN~\cite{zhu2017unpaired}, Transformer~\cite{vaswani2017attention}, and diffusion model~\cite{rombach2022high} for style transfer.
However, \lfl{directly applying these 2D methods} to 3D stylization will lead to issues like blurriness or view inconsistencies due to the inadequate consideration of 3D geometry.

\noindent\textbf{3D Gaussian Splatting. }
3D Gaussian Splatting (3DGS)~\cite{kerbl3Dgaussians} has emerged as \lfl{an} approach for real-time radiance field rendering and 3D scene reconstruction. 
Recent methods~\cite{chen2023gaussianeditor, GaussianEditor, gaussian_grouping, cen2023saga} enhance semantic understanding of 3D scenes and \lfl{enable} efficient text-based editing using pre-trained 2D models. \lfl{Additionally,} geometry deformation~\cite{gao2024mesh} and texture editing~\cite{wu2024deferredgs} have also been explored on 3DGS. Please refer to the survey~\cite{wu2024recent,chen2024survey} for more details. Despite these advancements, existing 3DGS works lack support for image-based 3D scene stylization that faithfully transfers detailed style features while offering flexible control. 
\begin{figure*}
    \centering
    \captionsetup{type=figure}
    \includegraphics[width=.98\linewidth]{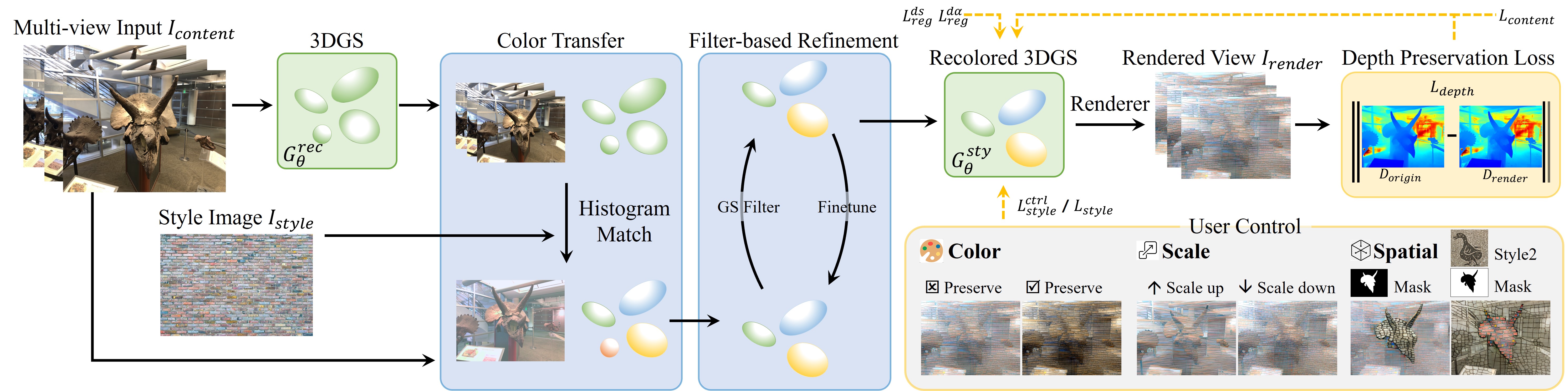}
    \vspace{-2mm}
    \captionof{figure}{\textbf{\name Pipeline.} We first reconstruct a photo-realistic 3DGS $G_{\theta}^{rec}$ from multi-view input. Following this, color matching with the style image is performed, accompanied by the filter-based refinement to preemptively address potential artifacts. 
    \lfl{During} optimization, we employ multiple loss terms to capture detailed local style structures and preserve geometric attributes. Users can flexibly control color, scale, and spatial attributes during stylization through customizable loss terms. Once this stylization is done, we can obtain consistent free-viewpoint stylized renderings. }
    \vspace{-2mm}
    \label{fig:pipeline}
\end{figure*}

\noindent\textbf{3D Style Transfer. }
With the increasing demand for 3D content, neural style transfer has been expanded to various 3D representations. Stylization on meshes often utilizes differential rendering to transfer style from rendered images to 3D meshes, enabling geometric or texture transfer~\cite{yin20213dstylenet, Kato_2018_CVPR, Michel_2022_CVPR}. Other works, using point clouds as the 3D proxy, ensure 3D consistency when stylizing novel views. LSNV~\cite{huang2021learning} employs featurized 3D point clouds modulated with the style image, followed by a 2D CNN renderer to generate stylized renderings. 
However, explicit methods' performance is constrained by the quality of geometric reconstructions, often leading to noticeable artifacts in complex real-world scenes.

Implicit methods, such as NeRF~\cite{mildenhall2021nerf}, have gained considerable attention for their enhanced capacity to represent complex scenes. Many NeRF-based stylization works incorporate image style transfer losses~\cite{gatys2015neural, zhang2022arf} during training or adopt a mutually learned image stylization network~\cite{huang2022stylizednerf,nguyen2022snerf} to optimize color-related parameters given a reference style image. Approaches like \cite{wang2023nerf, xu2023desrf} support both appearance and geometric stylization to mimic the reference style, achieving consistent results in novel-view stylization. However, these methods involve time-consuming optimization and exhibit slow rendering due to expensive random sampling in volume rendering. They also lack user-level flexible and accurate perceptual control for stylization. 

While controlling perceptual factors, such as color, stroke size, and spatial aspects, have been extensively explored in image style transfer \cite{gatys2017controlling, jing2018stroke, castillo2017zorn}, 
\lfl{they remain underdeveloped for 3D stylization.}
\cite{kumar2023s2rf, Zhang_2023_CVPR} establish semantic correspondence in transferring style across the entire stylized scene, \lfl{but they primarily focus on spatial control and lack the capability for} users to interactively specify arbitrary regions. 
ARF-plus~\cite{li2023arf} introduces more perceptual controllability into the stylization of radiance fields, yet the demand for enhanced flexibility and personalized, diverse characteristics in 3D stylization remains unmet.

In this work, leveraging the 3DGS representation, we achieve rapid stylization within a minute of training, ensuring real-time rendering capabilities. \zdx{Recent arXiv works~\cite{liu2024stylegaussian, saroha2024gaussian} on 3DGS-based style transfer exhibit limitations in maintaining intricate style details and coherence. \cite{liu2024stylegaussian} is feedforward-based and struggles to capture locally coherent style features and patterns with AdaIN~\cite{karras2019style}, while \cite{saroha2024gaussian} simply adopts classic style loss to optimize the color of Gaussians.}
Our method not only effectively captures distinctive details from the style image and preserves recognizable scene content with fidelity, but also empowers users with perceptual control over color, scale, and spatial factors for customized stylization.

\section{Method}
\begin{figure}[t]
    \centering
    \captionsetup{type=figure}
    \includegraphics[width=0.98\linewidth]{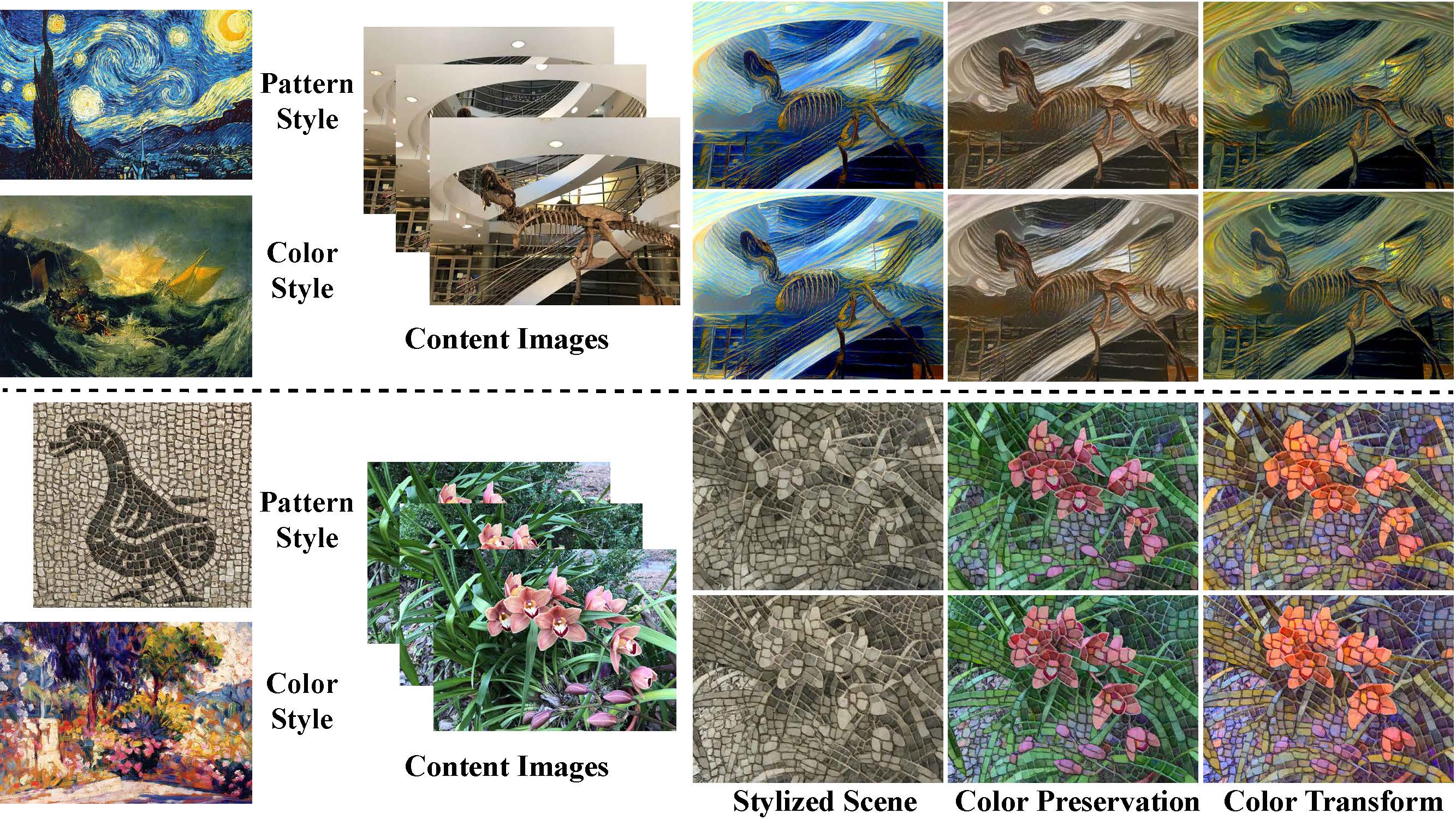}
    \vspace{-2mm}
    \captionof{figure}{\textbf{Color Control Results.} Our approach facilitates versatile color management in stylized outputs, allowing users to retain the scene's original hues or apply distinct color schemes from alternative style images. Users can choose to transfer the entire style, only the pattern style, or a mix of arbitrary patterns and color styles.}
    \vspace{-3mm}
    \label{fig:color_crtl}
\end{figure}
\zdxn{Our method enables efficient 3D style transfer by optimizing 3DGS representations~\cite{kerbl3Dgaussians} to transform captured scenes into consistently stylized outputs using arbitrary 2D style images. The pipeline comprises two sequential optimization stages: First, global scene recoloring aligns color statistics with the style image while filtering reconstruction floaters through Gaussian refinement. Second, local texture detailing employs nearest-neighbor feature matching to transfer intricate artistic patterns, regularized by depth preservation to maintain geometric integrity. This framework provides flexible control over color, scale, and spatial attributes (Sec.~\ref{sec:control}), enabling customizable artistic expression and consistent free-view rendering as demonstrated in Fig.~\ref{fig:pipeline}.}

\subsection{Preliminaries: 3D Gaussian Splatting} \label{sec:3dgs}
Gaussian Splatting \cite{kerbl3Dgaussians} encapsulates 3D scene information using a \lfl{set} of 3D colored Gaussians. This technique exhibits rapid inference speeds and exceptional reconstruction quality compared to NeRF. 
To represent the scene, each Gaussian is described by a centroid \zdxn{$x\in \mathbb{R}^3$}, a 3D vector $s \in \mathbb{R}^3$ for scaling, and a quaternion $q \in \mathbb{R}^4$ for rotation. Additionally, an opacity value $\alpha \in \mathbb{R}$ and a color vector $c$ represented in the coefficients of a spherical harmonic (SH) function of degree 3 are used for fast alpha-blending during rendering.
These trainable parameters are collectively symbolized by
$G_{\theta_i}$, where \zdxn{$G_{\theta_i} = \{x_i, s_i, q_i, \alpha_i, c_i\}$}, representing the parameters for the $i$-th Gaussian. To visualize the 3D Gaussians and supervise their optimization, 3DGS projects them onto the 2D image plane. The implementation leverages differentiable rendering and gradient-based optimization on each pixel for the involved Gaussians. The pixel color $c^{\alpha}$ is determined by blending the colors $c_i$ of those ordered Gaussians that overlap the pixel. This process can be formulated as:
\begin{equation}
    c^{\alpha}=\sum_{i\in N}T_i \alpha_i c_i
\end{equation}
where $T_i$ is the accumulated transmittance and $\alpha_i$ is the alpha-compositing weight for the $i$-th Gaussian. 

\subsection{Style Transfer to 3D Gaussian Splatting} \label{sec:style_transfer}
Given a set of content images $\mathcal{I}_{content}$ that are captured from different viewpoints in the same scene, we first reconstruct its 3DGS model $G_{\theta}^{rec}$. Our goal is to transform $G_{\theta}^{rec}$ to a stylized 3DGS model $G_{\theta}^{sty}$ that matches the detailed style features of a given 2D style image $\mathcal{I}_{style}$ while maintaining the content of the original scene.

\noindent\textbf{Color Transfer.} We first transfer the color distribution of the style image to the 3DGS $G_{\theta}^{rec}$, enhancing the alignment of hues with the style image. We will introduce how we enable a flexible color control later. Drawing inspiration from \cite{gatys2017controlling}, we employ a linear transformation of colors in RGB space, followed by a color histogram matching procedure between the style images and the views of the training set. Let ${p_s^i}$ represent the set of all pixels in the style image, and ${p_c^i}$ denote the set of all pixels in the content images to be recolored. We then solve the following linear transformation to align the mean ($\mu_{*}$) and covariance ($\Sigma_{*}$) of color distributions between the content image set and the style image: 
\begin{equation}
\begin{aligned}
    p_c^{re}=\mathbf{A}p_c+b, \, c^{re}=\mathbf{A}c+b \\  s.t.\, \mu_{p_c^{re}}=\mu_{p_s}, \Sigma_{p_c^{re}}=\Sigma_{p_s} 
\end{aligned}
\end{equation}
where $\mathbf{A} \in \mathbb{R}^{3 \times 3}$ and $b \in \mathbb{R}^3$ are the solution to the linear transformation, $p_c^{re}$ is a recolored pixel of the content images. The color parameter $c$ of 3DGS $G_{\theta}^{rec}$ is transformed to the recolored version $c^{re}$. Please refer to the supplementary document for the mathematical derivation.

\noindent\textbf{Filter-based Refinement.} 
After the color \lfl{transfer} step,  
\lfl{the color attributes $c^{re}$ of 3DGS and the recolored rendering images $\mathcal{I}_{content}^{re}$}
are consistent with the palette of the style image. 
\zdxn{However, some floaters in the original 3DGS will also be colored 
during this process, significantly affecting the quality of the stylization, as shown in Fig.~\ref{fig:ablation}. 
To address this issue, we design a filter-based refinement process to exclude these floaters while ensuring reconstruction quality before stylization. We use the recolored content images $\mathcal{I}_{content}^{re}$ as supervision to fine-tune the 3DGS with the recolored color parameter $c^{re}$ and filter the 3D Gaussian floater at every certain number of fine-tuning iterations. The filtering is performed based on the scale and opacity of the 3D Gaussians. Specifically, Gaussians whose sizes are in the top k\% or whose opacities are in the lowest k\% will be filtered out. The threshold k\% remains constant during the fine-tuning process. By maintaining a fixed threshold, we ensure that the reconstructed scene does not change significantly, as the fine-tuning process is relatively short. }
During fine-tuning, we incorporate the reconstruction loss from \cite{kerbl3Dgaussians}:
\begin{equation}
    \mathcal{L}_{rec}= (1 - \lambda_{rec})\mathcal{L}_1(\mathcal{I}_{content}^{re}, \mathcal{I}_{render}) + \lambda_{rec} \mathcal{L}_{D-SSIM} 
\end{equation}

\noindent\textbf{Stylization.} We then employ an intuitive and effective strategy to learn 3D Gaussian stylization by leveraging features extracted by a pre-trained convolutional neural network (e.g., VGG~\cite{simonyan2014very}).
The 3D Gaussians \lfl{are optimized} with a set of loss functions between training views and the style image. 
To transfer detailed high-frequency style features from a 2D style image to a 3D scene, we exploit the nearest neighbor feature matching concept inspired by \cite{zhang2022arf, kolkin2022neural}, where we minimize the cosine distance between the feature map of rendered images and its nearest neighbor in the style feature map. We extract the VGG feature maps $F_{style}$, $F_{content}$, and $F_{render}$ for $\mathcal{I}_{style}$, $\mathcal{I}_{content}^{re}$, and the rendered images $\mathcal{I}_{render}$, respectively. The nearest neighbor feature match loss is formulated as:
\begin{equation}
\label{eq:nnfm}
    \mathcal{L}_{style}(F_{render}, F_{style})=\frac{1}{N} \sum_{i,j}^{N}D(F_{render}(i,j), F_{style}(i*,j*))
\end{equation} 
where, $(i*,j*) = \arg\min_{i',j'}D(F_{render}(i,j), F_{style}(i',j'))$, and $D(a, b)$ is the cosine distance between two feature vectors $a$ and $b$. To preserve the original content structure during the stylization, we additionally minimize the mean squared distance between the content feature $F_{content}$ and the stylized feature $F_{render}$:
\begin{equation}
    \mathcal{L}_{content}=\frac{1}{H\times W}||F_{content}-F_{render}||^2
\end{equation}
where $H$ and $W$ represent the size of rendered images. While the content loss $\mathcal{L}_{content}$ mitigates the risk of excessive stylization, optimizations applied to the geometric parameters of 3D Gaussians can still induce changes in scene geometry. 
To address this, we introduce a depth preservation loss to maintain general geometric consistency without relying on \lfl{any} additional depth estimator. 
Specifically, in the color transfer stage, we initially generate the original depth map $D_{origin}$ by employing the alpha-blending method of 3DGS. The alpha-blended depth $d^{\alpha}$ of each pixel is calculated as $d^{\alpha}=\sum_{i}T_i \alpha_i d_i$, where $d_i$ is the depth value for $i$-th Gaussian. During the stylization, we minimize the $L_2$ loss between the rendered depth map $D_{render}$ and the original one $D_{origin}$:
\begin{equation}
    \mathcal{L}_{depth} = \frac{1}{H\times W}||D_{origin}-D_{render}||^2
\end{equation}
We also add some regularization terms that perform on the changes of scale $\Delta s$ and opacity $\Delta \alpha$ in $G_{\theta}^{sty}$:
\begin{equation}
    \mathcal{L}_{reg}^{ds}=\frac{1}{M}||\Delta s||, \quad \mathcal{L}_{reg}^{d\alpha}=\frac{1}{M}||\Delta \alpha||
\end{equation}
Finally, a total variation loss $\mathcal{L}_{tv}$ is used to smooth the rendered images in the 2D domain. The total loss during the stylization phase is:
\begin{equation} \label{eq:total_loss}
\begin{aligned}
\mathcal{L}=\lambda_{sty}\mathcal{L}_{style}+\lambda_{con}\mathcal{L}_{content}+\lambda_{dep}\mathcal{L}_{depth}\\
 +\lambda_{sca}\mathcal{L}_{reg}^{ds}+\lambda_{opa}\mathcal{L}_{reg}^{d\alpha} + \lambda_{tv}\mathcal{L}_{tv}
\end{aligned}
\end{equation}
{where $\lambda_*$ is the corresponding loss weight.}

\begin{figure}[htbp]
    \centering
    \captionsetup{type=figure}
    \includegraphics[width=0.98\linewidth]{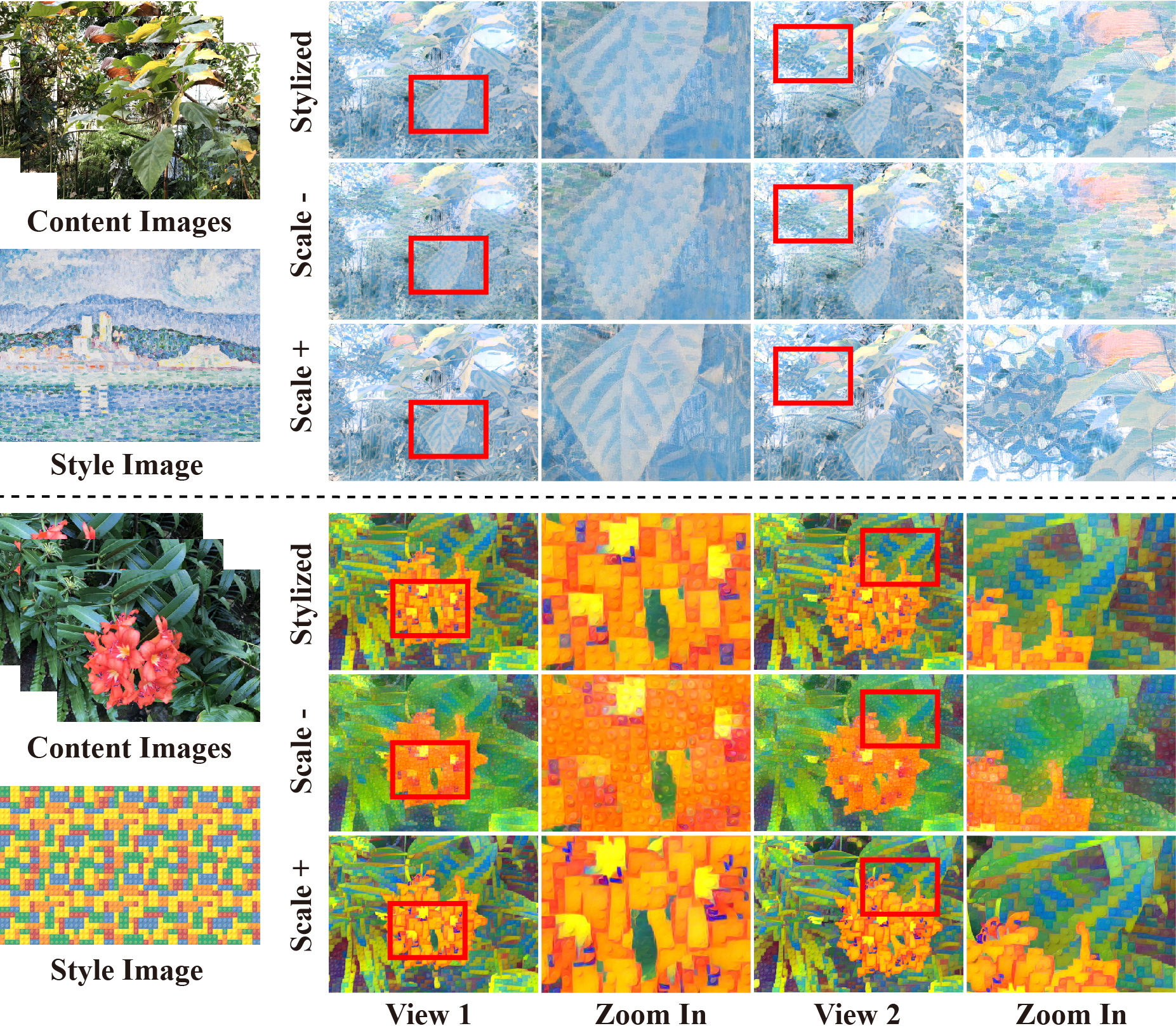}
    \vspace{-2mm}
    \captionof{figure}{\textbf{Scale Control Results.} Our method enables users to flexibly control the scale of basic style elements, such as adjusting the density of Lego blocks, as demonstrated in the example provided in the last row.}
    \vspace{-3mm}
    \label{fig:scale_crtl}
\end{figure} 

 \begin{figure}[t]
    \centering
    \captionsetup{type=figure}
    \includegraphics[width=0.98\linewidth]{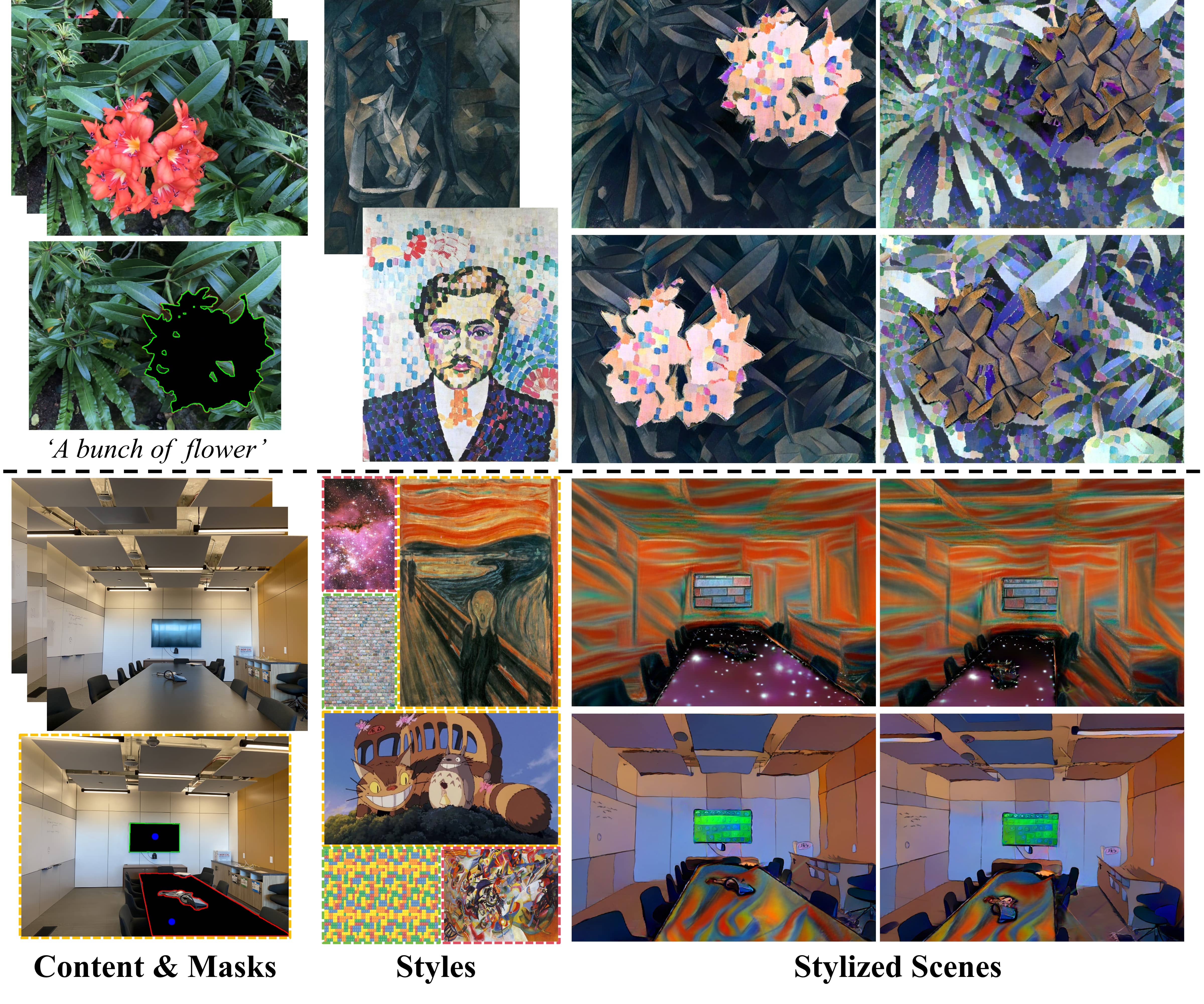}
    \vspace{-2mm}
    \captionof{figure}{\textbf{Spatial Control Results.} By specifying masks in the content images, users can transfer different styles to desired regions. Users can input text prompt or specify certain points (highlighted in blue) to generate region masks (depicted in black).}
    \vspace{-6mm}
    \label{fig:spatial_crtl}
\end{figure}

\begin{figure}[t]
    \centering
    \captionsetup{type=figure}
    \includegraphics[width=0.98\linewidth]{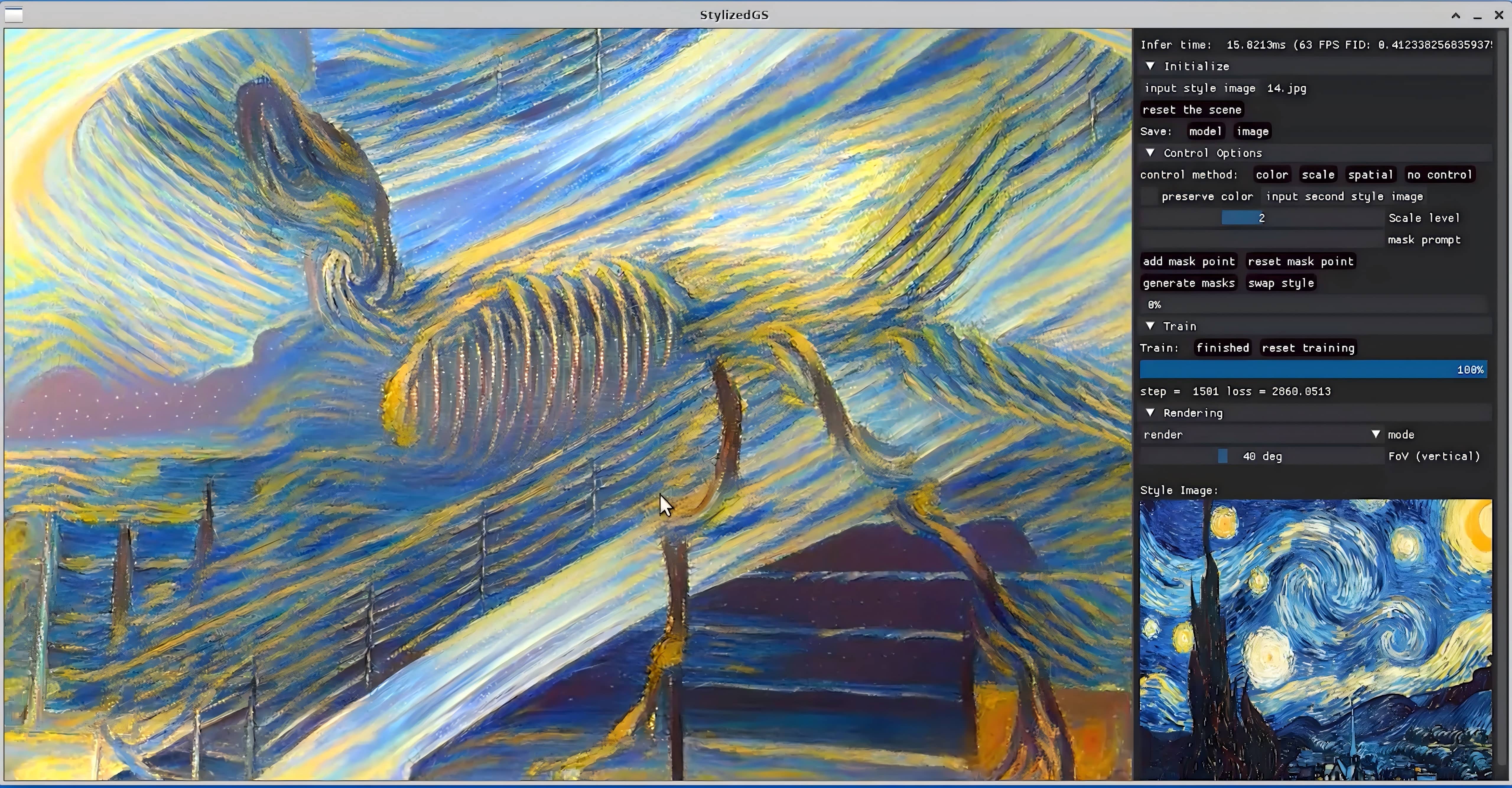}
    \vspace{-2mm}
    \captionof{figure}{\textbf{The User Interface of StylizedGS.}  \zdx{Users can upload style and multi-view scene images via the right control panel, and the stylization results are displayed in the left panel in real time after optimization. The interface supports view control, color control, scale control, and spatial control with point-based and language-based interactions.}}
    \vspace{-6mm}
    \label{fig:ui}
\end{figure}

\subsection{Perceptual Control} \label{sec:control}
\begin{figure*}[htbp]
    \centering
    \captionsetup{type=figure}
    \includegraphics[width=\linewidth]{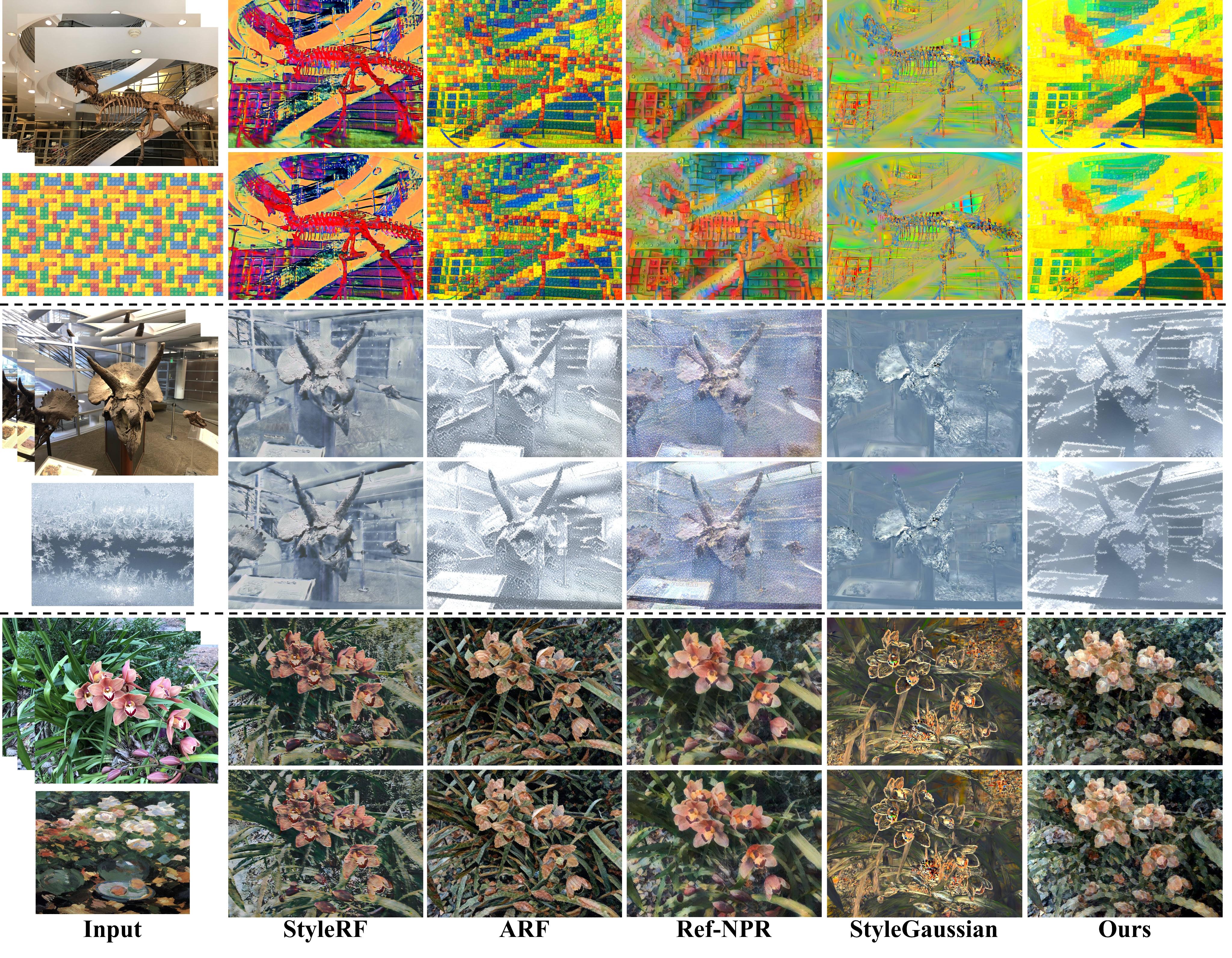}
    \vspace{-6mm}
    \captionof{figure}{\textbf{Qualitative comparisons with baselines on LLFF dataset.} Compared to other methods, our method excels in learning intricate and accurate geometric patterns while effectively preserving the semantic content of the original scene. (\textbf{Please zoom in for better view.})}
    \label{fig:comp_llff}
\end{figure*}
Building on the previous components of our approach, we can achieve both rapid and intricate style transfer while preserving the original content. Here, we address the challenge users face in customizing the stylization process. Inspired by \cite{gatys2017controlling}, we propose a series of strategies and loss functions for 3DGS stylization to control various perceptual factors, including color, scale, and spatial areas. \yyj{The stylization loss will be replaced with a different form, namely $L_{style}^{color}$, $L_{style}^{scale}$ or $L_{style}^{spatial}$ defined below.}

\noindent\textbf{Color Control.} Color information within an image is a crucial perceptual aspect of its style. Unlike other stylistic elements such as brush strokes or dominant geometric shapes, color is largely independent and stands out as a distinct characteristic. As illustrated in Fig.~\ref{fig:color_crtl}, users may seek to preserve the original color scheme by transferring only the pattern style or by combining the pattern style of one image with the color style of another. To support this flexibility, our method offers independent control over color during stylization.

\zdxn{
We achieve this by pre-coloring the 3D scene to match user-specified hues, followed by style transfer in the YIQ color space, which decouples luminance and chrominance. Specifically, our color style loss $\mathcal{L}{\text{style}}^{\text{color}}$ (Eq.~\ref{eq:color}) is computed using only the luminance channel of the rendered view $\mathcal{I}{\text{render}}^Y$ and the style image $\mathcal{I}{\text{style}}^Y$. Meanwhile, RGB channels are retained for the content loss $\mathcal{L}{\text{content}}$, allowing for structural fidelity to be maintained during stylization.}
\begin{equation} \label{eq:color}
    \mathcal{L}_{style}^{color}=\mathcal{L}_{style}(F(\mathcal{I}_{render}^Y), F(\mathcal{I}_{style}^Y))
\end{equation}

\noindent\textbf{Scale Control.} 
The scale-related style features, such as the thickness of brushstrokes or the density of the basic grains (as depicted in Fig.~\ref{fig:scale_crtl}), are foundational style elements and play a vital role in defining visual aesthetics. \yyj{Here, the `scale' we mention refers to the style pattern size.} The stroke size, as a typical example of scale features, is influenced by two factors: the receptive field of the convolution in the VGG network and the size of the style image input to the VGG network, as identified by \cite{jing2018stroke}. Our exploration reveals that simply adjusting the receptive field is efficient for achieving scale-controllable stylization in 3DGS. We modulate the size of the receptive field by manipulating the selection of layers in the VGG network and adjusting the weights for each layer during the computation of the scale style loss $\mathcal{L}_{style}^{scale}$. As shown in Eq.~\ref{eq:scale}, the style losses from different layer blocks are combined as $\mathcal{L}_{style}^{scale}$:
\begin{equation} \label{eq:scale}
    \mathcal{L}_{style}^{scale}=\sum_{l\in b_s}w_l\cdot \mathcal{L}_{style}(F_l(\mathcal{I}_{render}), F_l(\mathcal{I}_{style}))
\end{equation}
Here, $F_l$ represents the feature map at the $l$-th layer within the $s$-th VGG-16 block $b_s$, and $w_l$ controls the corresponding weights at the $l$-th layer.

\noindent\textbf{Spatial Control.} Traditional 3D style transfer methods typically apply a single style uniformly across the entire scene, resulting in uniform stylization throughout the composition. However, there are scenarios where it becomes necessary to transfer distinct styles to different areas or objects. Users may also wish to specify region-to-region constraints for their preferred style distribution within a scene, as illustrated in Fig.\ref{fig:spatial_crtl}. \yyj{We provide point-based interaction using SAM~\cite{kirillov2023segany} and a proposed mask-tracking strategy, and a language-based approach using the LangSAM method~\cite{langsam_github} to generate masks across different views. Further details can be found in the supplementary document.}
Similarly, users can also specify certain areas in the style image to delineate style regions for partial style transfer.

To provide spatial guidance during the stylization process, we introduce a spatial style loss, denoted as $\mathcal{L}_{style}^{spatial}$. Assuming the user specifies $r$ regions in a single view of a scene to be matched with $r$ style regions, the spatial loss $\mathcal{L}_{style}^{spatial}$ is formulated as follows:
\begin{equation} \label{eq:spatial}
     \mathcal{L}_{style}^{spatial}=\sum_r w_r\cdot \mathcal{L}_{style}(F(M_r^c \circ \mathcal{I}_{render}^r), F(M_r^s \circ \mathcal{I}_{style}^r))
\end{equation}
Here, $M_r^c$ and $M_r^s$ represent the binary spatial masks on the rendered views and the $r$-th style image, respectively. 
Additionally, during the color transfer step, we transform the color information in the region of the style image to the corresponding masked area of the content image to improve color correspondence.

\subsection{User Interface}
\zdx{
We develop an interactive user interface integrated with our proposed method to facilitate convenient image-based scene stylization and various perceptual controls, as illustrated in Fig.~\ref{fig:ui}.  Users can upload a style image and multi-view scene images through the right control panel to generate stylized scenes. After the optimization, our interface displays the stylized scenes in the left window which can be viewed in real time. Additionally, the system supports interactive controls and various render modes by adjusting options in the control panel. Users can change the views by dragging the display window. For spatial control, we have implemented both point-based and language-based interactions to flexibly generate masks. Furthermore, users can customize the stylized scenes by adjusting the optimization steps and loss functions, combining different perceptual controls and style images, thereby supporting detailed and expressive stylization. We also provide a video of manipulating the interactive interface in the supplementary material.
}

\section{Experiments} \label{sec:exp}

 \begin{figure*}[ht]
    \centering
    \includegraphics[width=0.9\linewidth]{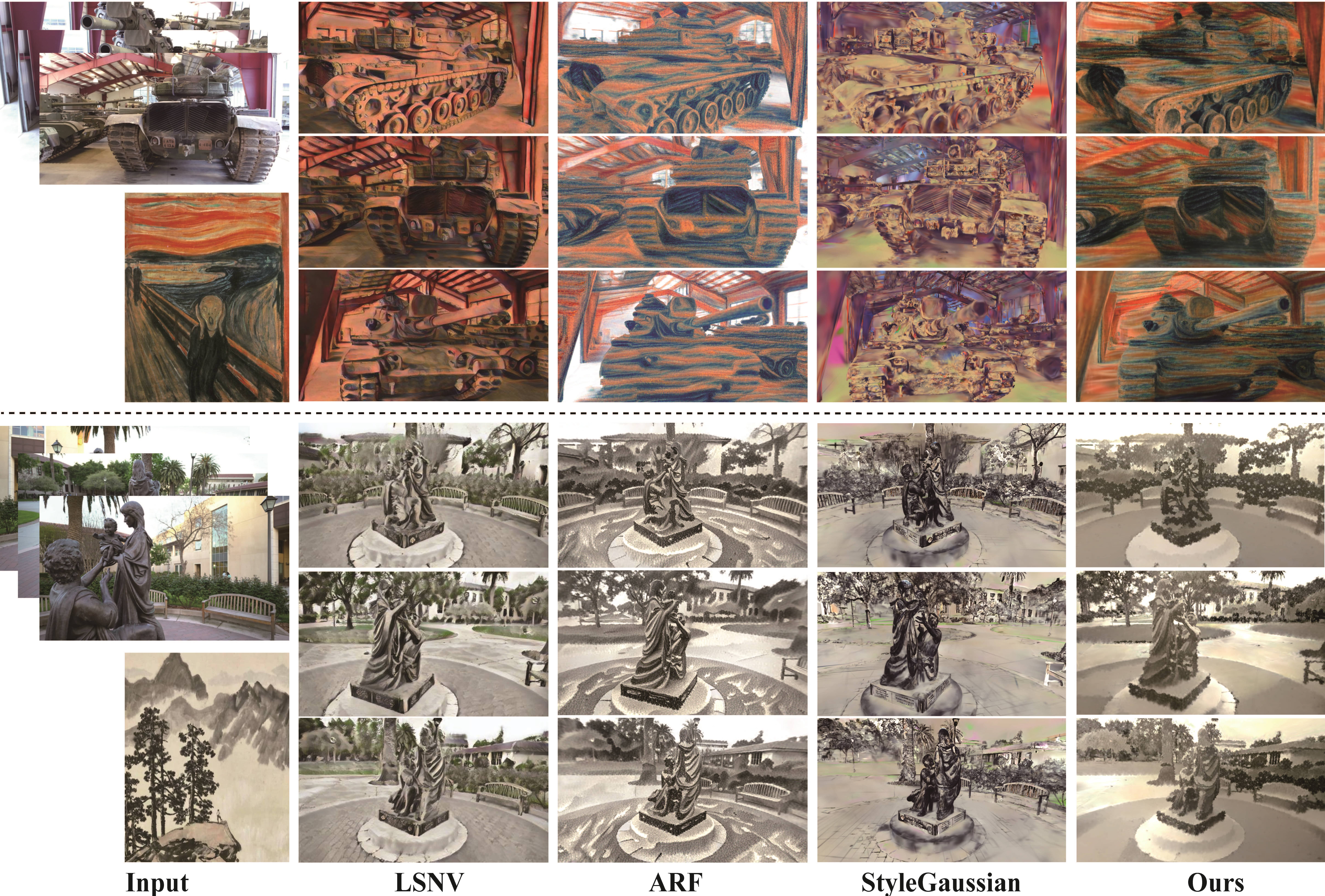}
    \vspace{-3mm}
    \caption{\textbf{Qualitative comparisons with the baseline methods on T\&T datasets.} It can be seen that our method has better results which faithfully capture both the color styles and pattern styles across different views. (\textbf{Please zoom in for better view.})}
    \vspace{-2mm}
    \label{fig:comp_tnt}
\end{figure*}

\begin{figure*}[htbp]
    \centering
    \includegraphics[width=0.99\linewidth]{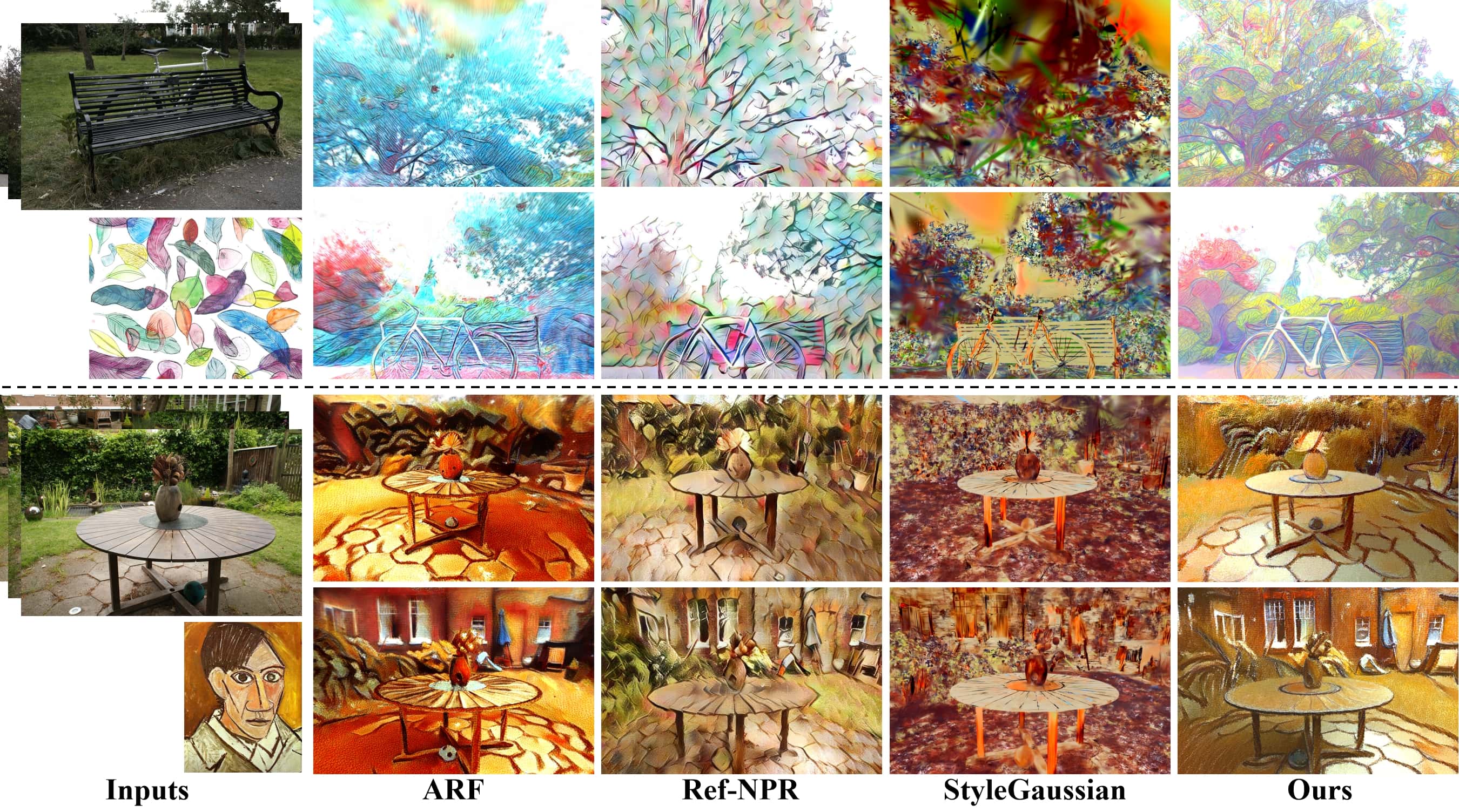}
    \caption{\zdxn{\textbf{Qualitative comparisons with the baseline methods on MipNeRF-360 datasets.} (\textbf{Please zoom in for better view.})}}
    \label{fig:comp_mipnerf}
\end{figure*}
We include the implementation details, especially the training settings in our supplementary document.

\textbf{Datasets.} 
\zdxn{We conduct extensive experiments on multiple real-world scenes in LLFF~\cite{mildenhall2019llff} which contains forward-facing scenes, Tanks \& Temples (T\&T) ~\cite{knapitsch2017tanks} and MipNeRF-360 \cite{barron2022mipnerf360} dataset which includes unbounded 360$^\circ$ large outdoor scenes. Additionally, we demonstrate the generality of our method by presenting stylization results on the Deep Blending~\cite{DeepBlending2018} and Instruct-NeRF2NeRF~\cite{instructnerf2023} datasets.}
Furthermore, our experiments involve a diverse set of style images from \cite{zhang2022arf} and WikiArt~\cite{artgan2018}, allowing us to assess our method's capability to handle a wide range of stylistic exemplars.

\noindent\textbf{Baselines.}
We compare our method with the state-of-the-art 3D stylization methods, including LSNV~\cite{huang2021learning}, ARF~\cite{zhang2022arf}, StyleRF~\cite{liu2023stylerf}, Ref-NPR~\cite{Zhang_2023_CVPR} and StyleGaussian~\cite{liu2024stylegaussian}, which are based on different 3D representations. Specifically, LSNV is based on point cloud, ARF, StyleRF and Ref-NPR are based on NeRF, and StyleGaussian is based on 3DGS. LSNV, StyleRF and StyleGaussian are feed-forward-based, while others are optimization-based.
For all methods, we use their released code and pre-trained models. As Ref-NPR is a reference-based method, we use AdaIN~\cite{karras2019style} to obtain a stylized reference view according to its paper. 

\subsection{Qualitative Comparisons}
\begin{figure}
    \centering
    \captionsetup{type=figure}
    \includegraphics[width=0.98\linewidth]{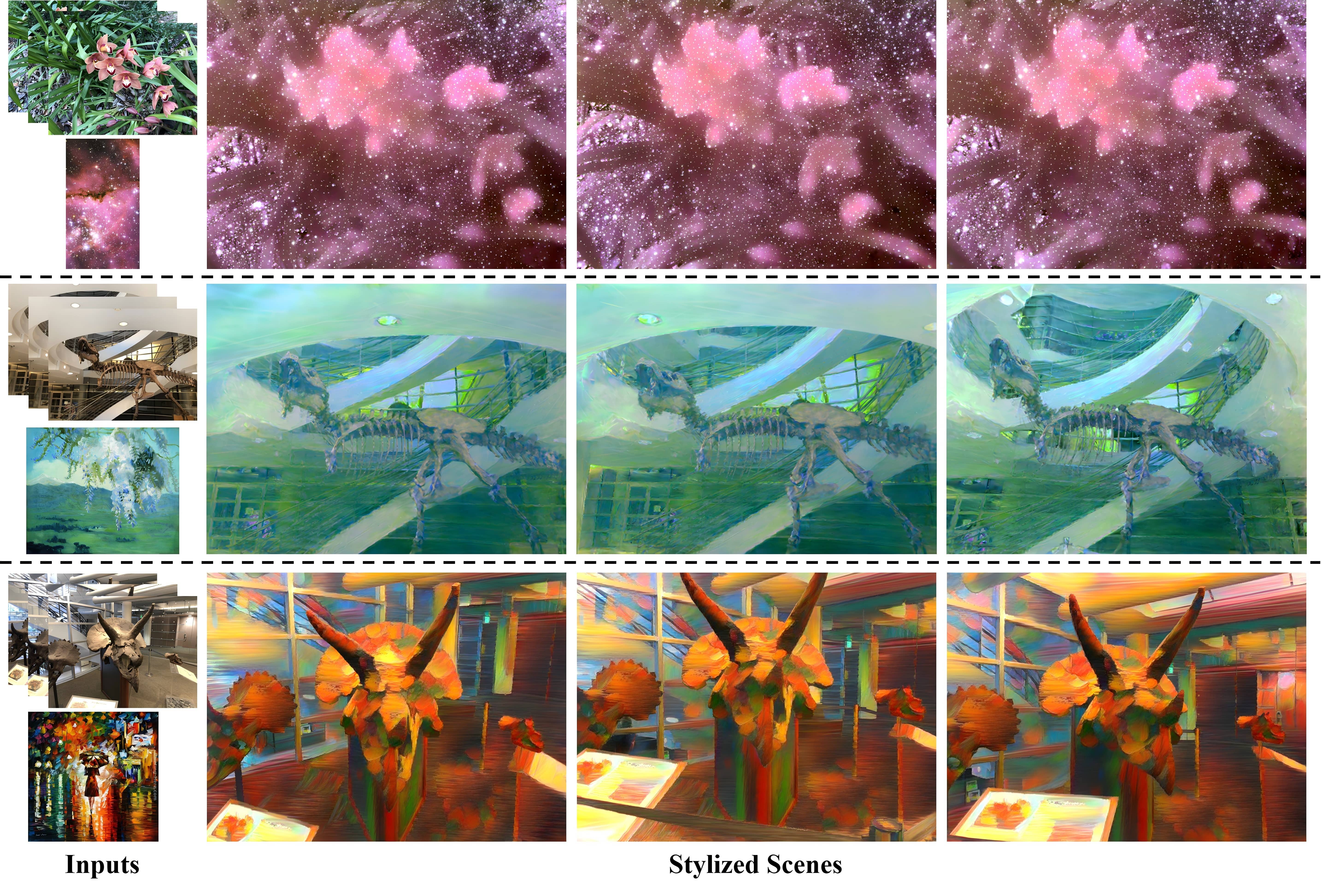}
    \captionof{figure}{\textbf{Stylization results on LLFF dataset.} 
Our method demonstrates robust performance, effectively adapting to various styles and diverse scenes.}
    \label{fig:more_llff}
    \vspace{-6mm}
\end{figure}

\begin{figure}
    \centering
    \captionsetup{type=figure}
    \includegraphics[width=\linewidth]{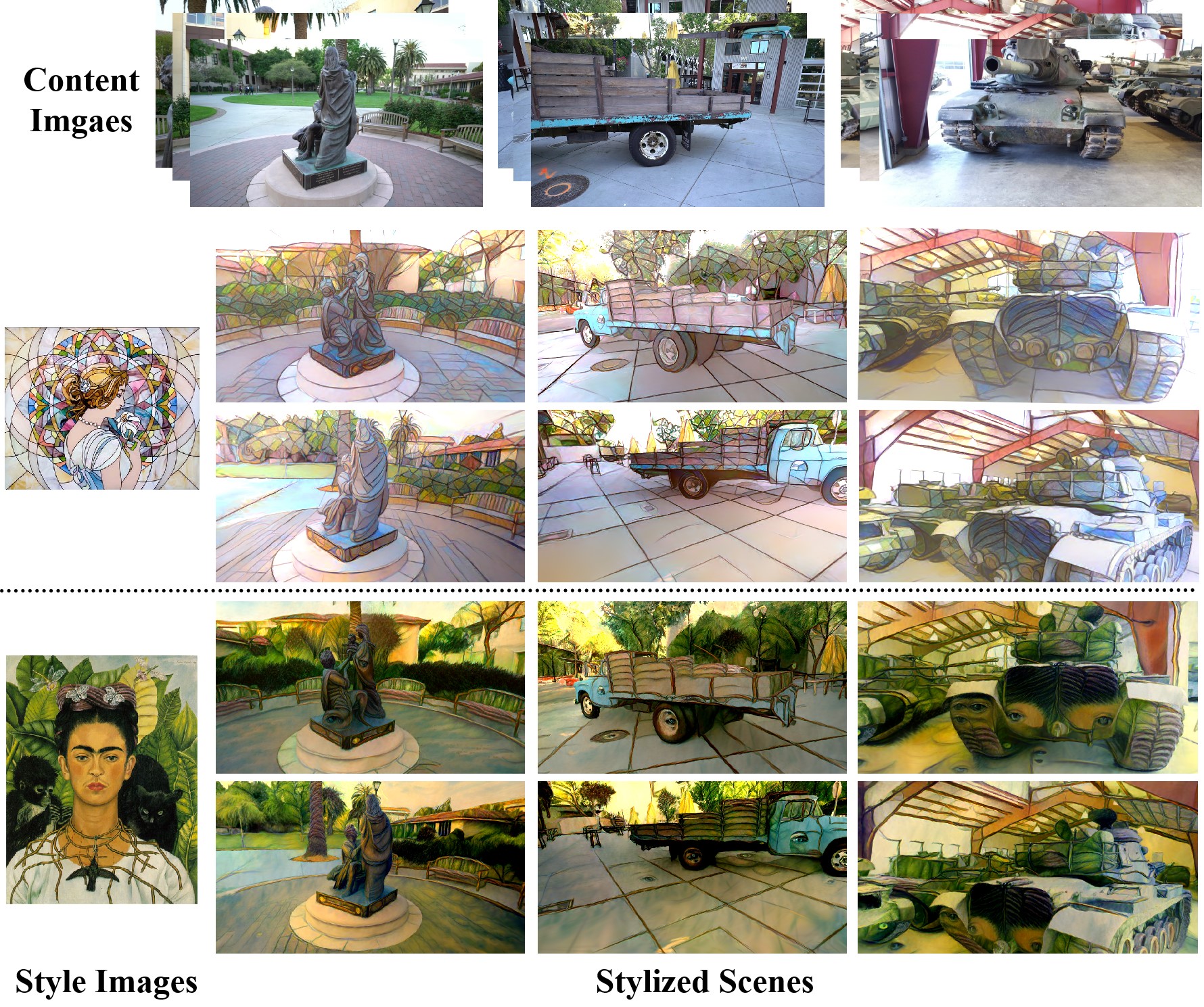}
    \vspace{-6mm}
    \captionof{figure}{\textbf{Stylization results on T\&T dataset.} It can be seen that our method faithfully captures both the color styles and pattern styles across different views.}
    \label{fig:stylization_matrix}
\end{figure}

\begin{figure}
    \centering
    \captionsetup{type=figure}
    \includegraphics[width=\linewidth]{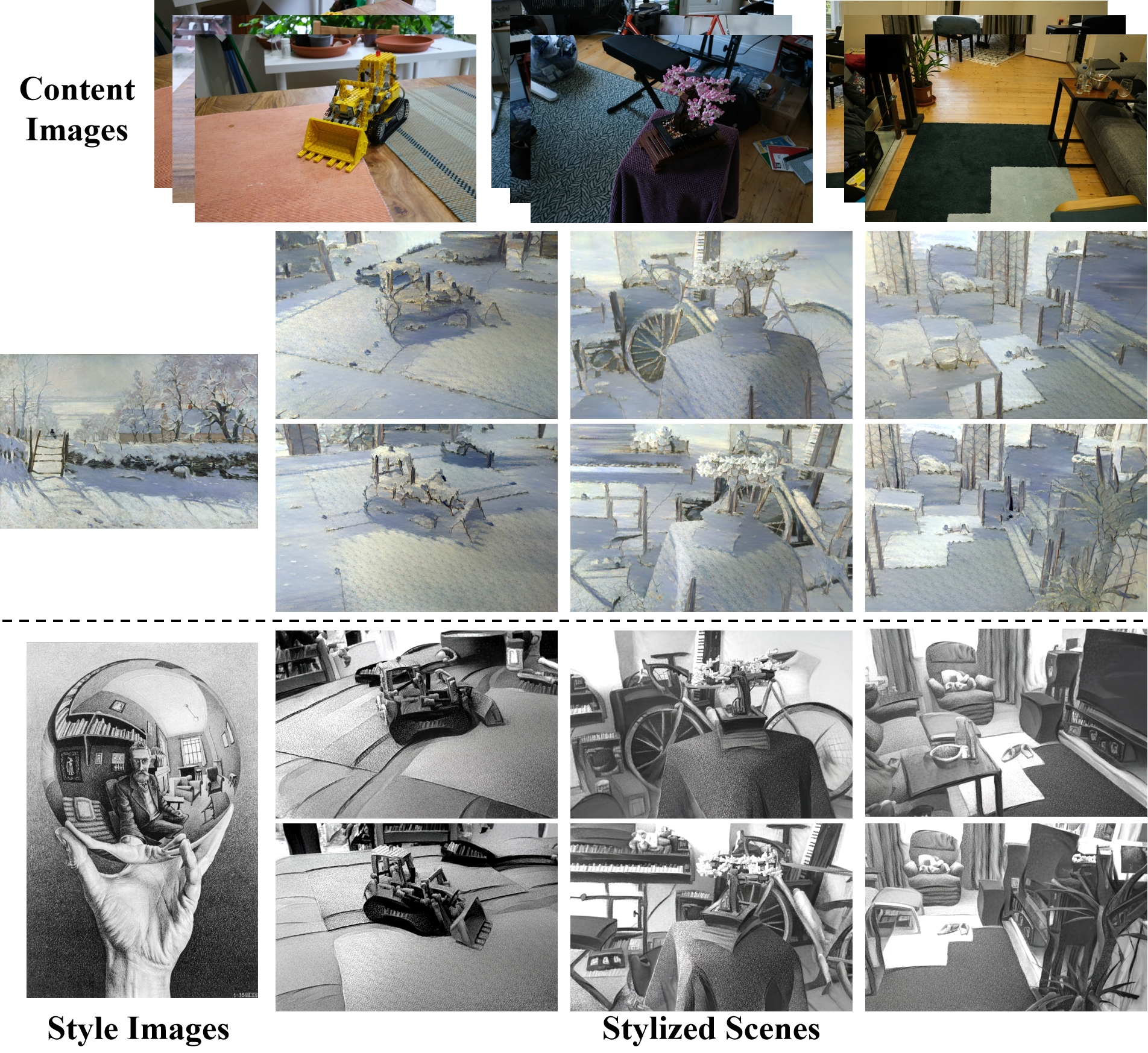}
    \vspace{-4mm}
    \captionof{figure}{\textbf{Stylization results on MipNeRF-360 dataset.}}
    \label{fig:stylization_mipnerf}
\end{figure}
We show visual comparisons on the LLFF dataset in Fig.~\ref{fig:comp_llff}. By jointly optimizing the opacity and color components, our method performs better in learning intricate geometric patterns and meso-structures. Specifically, for the scene of ``trex'' with a Lego style, other methods capture only the color style or a basic square pattern, whereas our approach successfully illustrates the detailed structure of the Lego blocks, clearly depicting even the small cylindrical patterns. This is also evident in the scene of "horns" with a snow style, where our method produces visually more pleasing results that faithfully match the style of the hexagonal snowflake structure. Additionally, our method preserves more semantic information in the stylized scene by utilizing the depth map to maintain spatial distribution in the image.

\begin{figure}
    \centering
    \captionsetup{type=figure}
    \includegraphics[width=\linewidth]{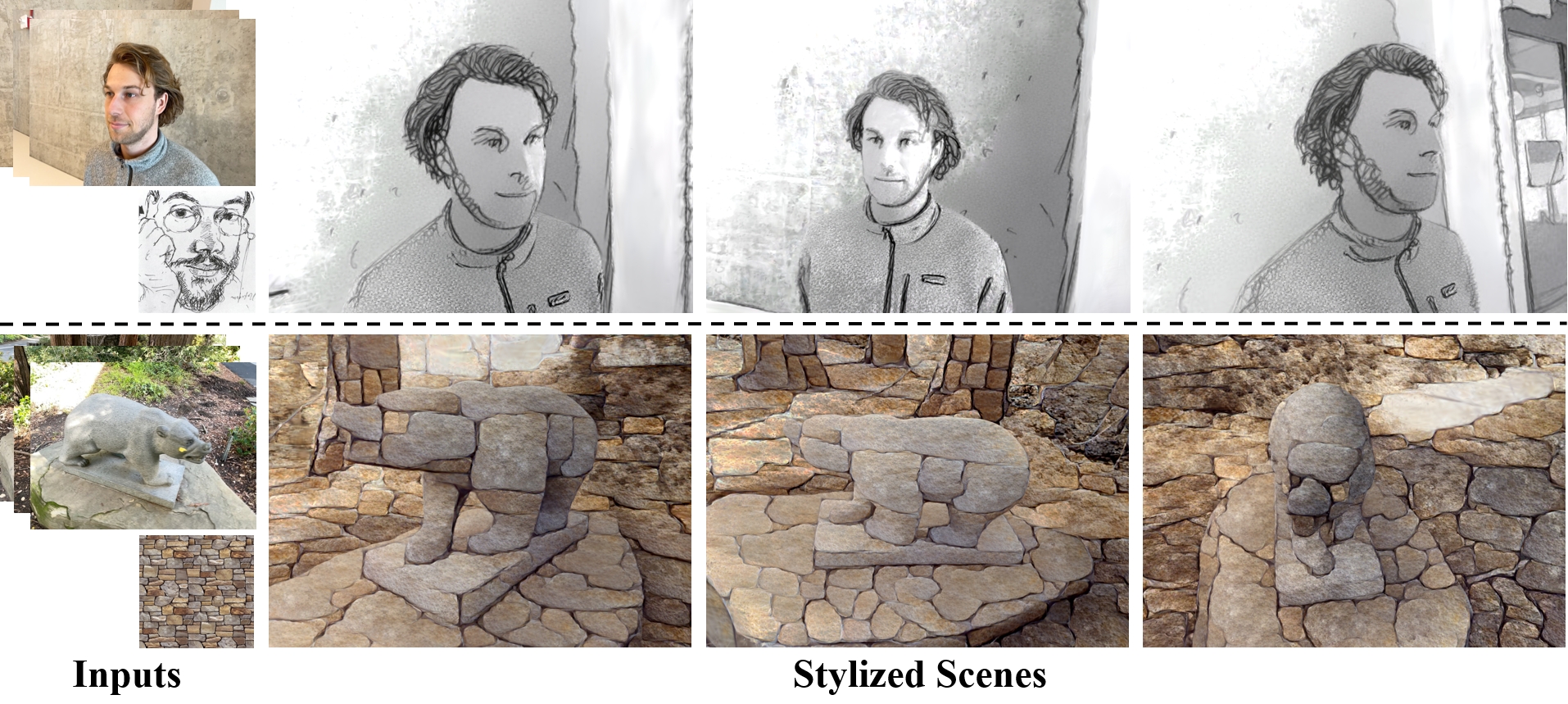}
    \vspace{-4mm}
    \captionof{figure}{\textbf{Stylization results on Instruct-NeRF2NeRF dataset.}}
    \label{fig:stylization_in2n}
\end{figure}

\begin{figure}
    \centering
    \captionsetup{type=figure}
    \includegraphics[width=\linewidth]{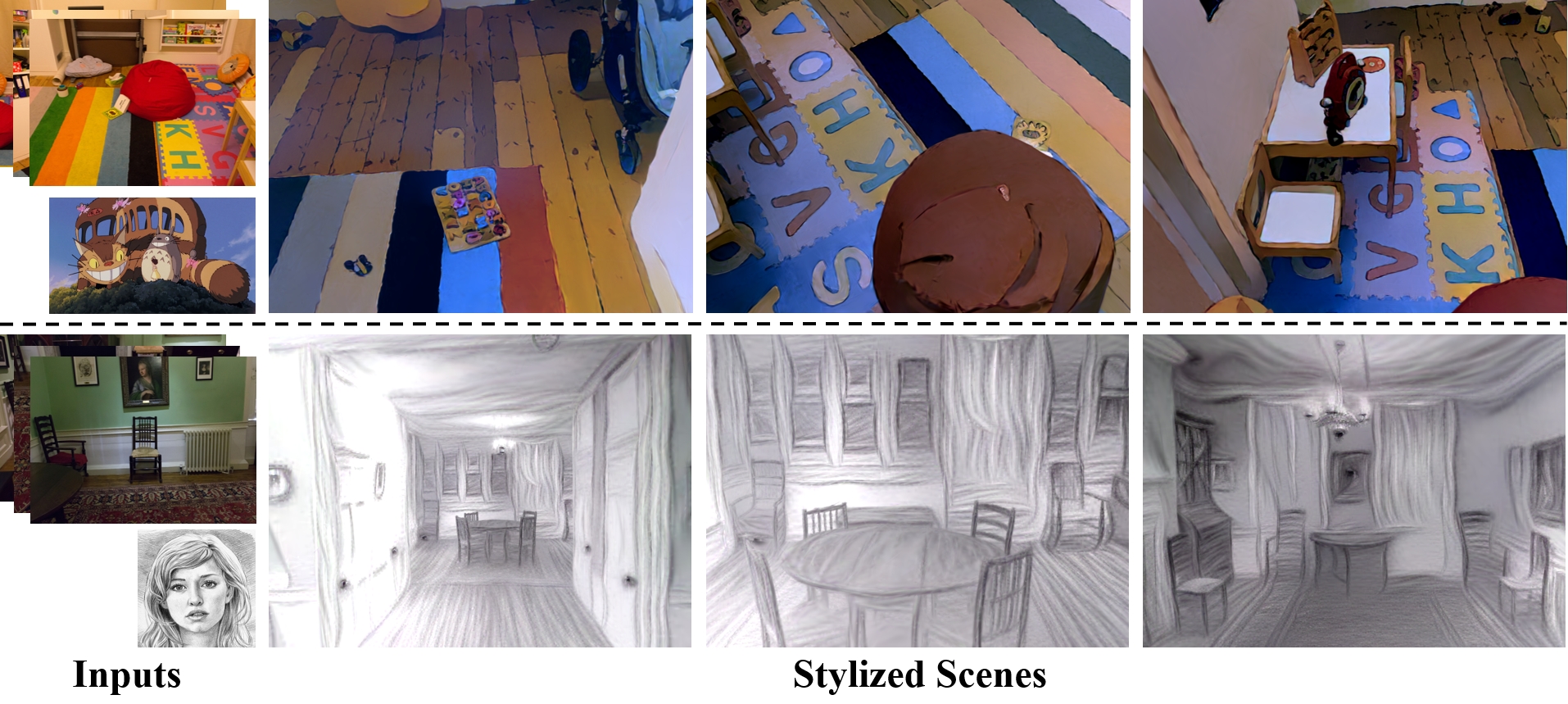}
    \vspace{-4mm}
    \captionof{figure}{\textbf{Stylization results on Deep Blending dataset.}}
    \label{fig:stylization_db}
\end{figure}

\zdxn{Fig.~\ref{fig:comp_tnt} and Fig.~\ref{fig:comp_mipnerf} presents comparison results on the T\&T and MipNeRF-360 dataset. It can be seen that our method achieves a better style match with the style image compared to the others. For instance, LSNV tends to produce overly smoothed results and lacks intricate structures, while ARF and Ref-NPR struggles to consistently transfer the style across the entire scene and exhibits deficiencies in color correspondence, such as the results of the second group. StyleGaussian does not faithfully transfer the style, resulting in blurry and noise results. In contrast, our method accurately captures both the color tones and brushstrokes throughout the entire scene, demonstrating superior style matching compared to the baselines. 

We present additional stylization results across different datasets: in Fig.~\ref{fig:more_llff}, Fig.~\ref{fig:stylization_matrix}, Fig.~\ref{fig:stylization_in2n} and Fig.~\ref{fig:stylization_db}. Please refer to our supplementary for more stylization results.}

\begin{table}[t]
\setlength\tabcolsep{4pt}
\caption{Quantitative comparisons on stylization under novel views. We report ArtFID, SSIM, DISTS, average training time (Train), and average rendering FPS (FPS) for our method and other baselines. `-' denotes that the method does not require individual stylization training.}
\vspace{-2mm}
\resizebox{\linewidth}{!}{
  \begin{tabular*}{\linewidth}{@{}cccccc@{}}
    \toprule
    Metrics & ArtFID($\downarrow$) & SSIM($\uparrow$) & DISTS($\downarrow$) & Train($\downarrow$) & FPS($\uparrow$) \\
    \midrule
    LSNV & 52.75 & 0.13 & 0.33 & - & 0.71\\
    StyleRF    & 40.61 & 0.41 & 0.30 & - & 5.8 \\
    ARF        & 35.73 & 0.29 & 0.31 & 2.25 min & 8.2 \\
    Ref-NPR    & 33.56 & 0.31 & 0.33 & 2.54 min & 7.3 \\
    StyleGaussian & 37.92 & 0.35 & 0.30 & - & 142\\
    Ours       & \textbf{28.29}& \textbf{0.55} & \textbf{0.26} & \textbf{0.87 min}& \textbf{153}\\
  \bottomrule
\end{tabular*}
}
\vspace{-4mm}
\label{tab:comp}
\end{table}

\begin{table}[htbp]
\setlength\tabcolsep{4pt}
\caption{\zdxn{Quantitative comparisons on stylization under novel views on MipNeRF-360 dataset. We report ArtFID, SSIM, DISTS and average rendering FPS (FPS) for our method and other baselines.}}
\centering
\vspace{-2mm}
  \begin{tabular}{@{}ccccc@{}}
    \toprule
    Metrics & ArtFID ($\downarrow$) & SSIM ($\uparrow$) & DISTS ($\downarrow$) & FPS($\uparrow$) \\
    \midrule
    ARF        & 34.79 & 0.25 & 0.35 & 7.6 \\
    Ref-NPR    & 35.16 & 0.27 & 0.33 & 5.3 \\
    StyleGaussian & 45.38 & 0.24 & 0.32 & 116\\
    Ours       & \textbf{29.12}& \textbf{0.40} & \textbf{0.28} & \textbf{137}\\
  \bottomrule
\end{tabular}
\vspace{-4mm}
\label{tab:comp_mipnerf}
\end{table}

\begin{table}[t]
\caption{\zdxn{\textbf{Quantitative results.} 
We evaluate the performance of our method against the state-of-the-art in terms of consistency, using LPIPS (\(\downarrow\))  and RMSE (\(\downarrow\)).}}
\centering

\begin{tabular}{@{\hspace{0.7em}}c@{\hspace{0.7em}}|@{\hspace{0.7em}}c@{\hspace{0.7em}}c@{\hspace{0.7em}}c@{\hspace{0.7em}}c@{\hspace{0.7em}}}
\toprule

Methods & 
\multicolumn{2}{c}{\begin{tabular}{@{}c@{}}Short-range\\Consistency\end{tabular}} & 
\multicolumn{2}{c@{\hspace{0.7em}}}{\begin{tabular}{@{}c@{}}Long-range\\Consistency\end{tabular}} \\

\midrule

{}  & \textit{LPIPS} & \textit{RMSE}  & \textit{LPIPS} & \textit{RMSE} \\

ARF &  0.036  &  0.051 &  0.185  &  0.128 \\
Ref-NPR &  0.035  & 0.052  &  0.173 &  0.124  \\
StyleGaussian &  0.031  & 0.048  &  0.159 &  0.117  \\
Ours &  \textbf{0.028}  & \textbf{0.044} & \textbf{0.112} &  \textbf{0.088}  \\
\bottomrule
\end{tabular}

\label{tab:comp_mv}
\end{table}

\subsection{Quantitative Comparisons}
\zdxn{%
To assess both fidelity and efficiency, we perform quantitative comparisons with existing methods. }
For the stylization metrics, we choose ArtFID~\cite{wright2022artfid} to measure the quality of stylization following \cite{huang2023quantart}. We also use SSIM~\cite{ssim} and DISTS~\cite{ding2020iqa} between the content and stylized images to measure the performance of detail preservation and structural similarity following \cite{an2021artflow, hong2021domain, yang2022industrial, kim2021deep, dist_cite}. As shown in Tab.~\ref{tab:comp} and Tab.~\ref{tab:comp_mipnerf}, our method excels in both stylized visual quality and content preservation. 
\zdxn{In terms of efficiency, our approach supports real-time free-viewpoint rendering and significantly reduces per-scene stylization time, achieving sub-minute training on a single RTX 4090 thanks to 3DGS. While StyleRF and StyleGaussian are feed-forward methods that require no per-scene training, they suffer from slower rendering and reduced stylization fidelity.

We perform quantitative comparisons on multi-view consistency in Tab. \ref{tab:comp_mv}, adopting established view-warping protocols \cite{chiang2022stylizing, fan2022unified, huang2021learning, liu2023stylerf, liu2024stylegaussian, dhiman2023corf}. Using optical flow \cite{teed2020raft} with softmax splatting \cite{niklaus2020softmax}, we warp stylized views between camera poses and compute masked RMSE and LPIPS \cite{zhang2018unreasonable} scores to quantify consistency. Our evaluation spans 5 unbounded scenes from T\&T and MipNeRF-360 datasets, employing 20 view pairs per scene at short-range (5-frame gap) and long-range (35-frame gap) intervals. Each pair is stylized with 10 distinct style images, yielding 200 comprehensive test cases. As evidenced in Tab. \ref{tab:comp_mv}, our method demonstrates superior consistency metrics, overcoming the view-dependent artifacts inherent in NeRF-based approaches and the feature decoding limitations of StyleGaussian's 3D CNN architecture. }

\begin{figure}[htbp]
    \centering
    \includegraphics[width=.98\linewidth]{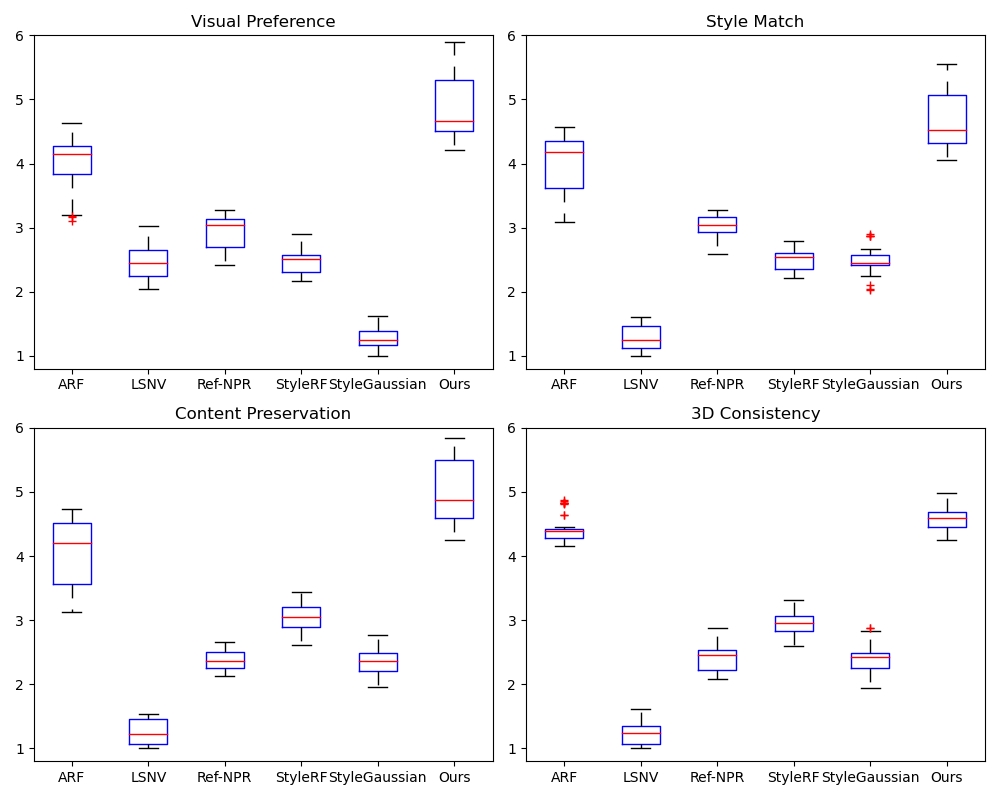}
    \vspace{-3mm}
    \caption{\zdxn{\textbf{User Study.} We record the user preference in the form of a boxplot. Our results obtain more preferences in visual preference, style match level, content preservation level, and 3D consistency quality than other state-of-the-art stylization methods.}}
    \label{fig:user_study_new}
\end{figure}

\begin{figure*}[htbp]
    \centering
    \captionsetup{type=figure}
    \includegraphics[width=\linewidth]{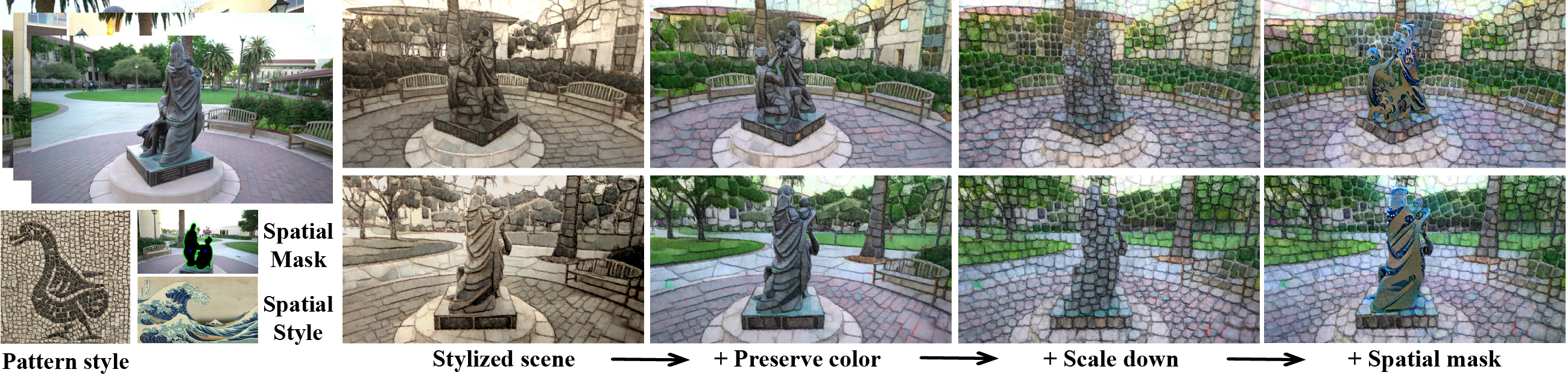}
    \vspace{-6mm}
    \captionof{figure}{\textbf{Sequential control results.} Given multiple control conditions, we can achieve a sequence of controllable stylization. We first show a stylization result and then, from left to right, progressively implement controls that preserve the color of the original scene, increase the scale of the pattern style, and adopt spatial control to apply two styles.}
    \label{fig:serial_ctrl}
\end{figure*}

\begin{table}[htbp]
\centering
\setlength\tabcolsep{4pt}
\caption{Quantitative ablation study results. We report ArtFID and SSIM for our method and other ablations over 50 randomly chosen stylized cases.}
\vspace{-2mm}
  \begin{tabular*}{\linewidth}{@{}cccccc@{}}
    \toprule
    Metrics & \scriptsize w/o  recolor & \scriptsize w/o density & \scriptsize w/o depth & \scriptsize w/o GS filter & \scriptsize Ours \\
    \midrule
    ArtFID ($\downarrow$) & 31.57 & 38.54 & 31.20 & 30.02 & \textbf{28.29}\\
    SSIM ($\uparrow$) & 0.54 & 0.53 & 0.35 & 0.39 & \textbf{0.55} \\

  \bottomrule
\end{tabular*}
\vspace{-4mm}
\label{tab:ablation}
\end{table}

\begin{figure}[htbp]
    \centering
    \includegraphics[width=\linewidth]{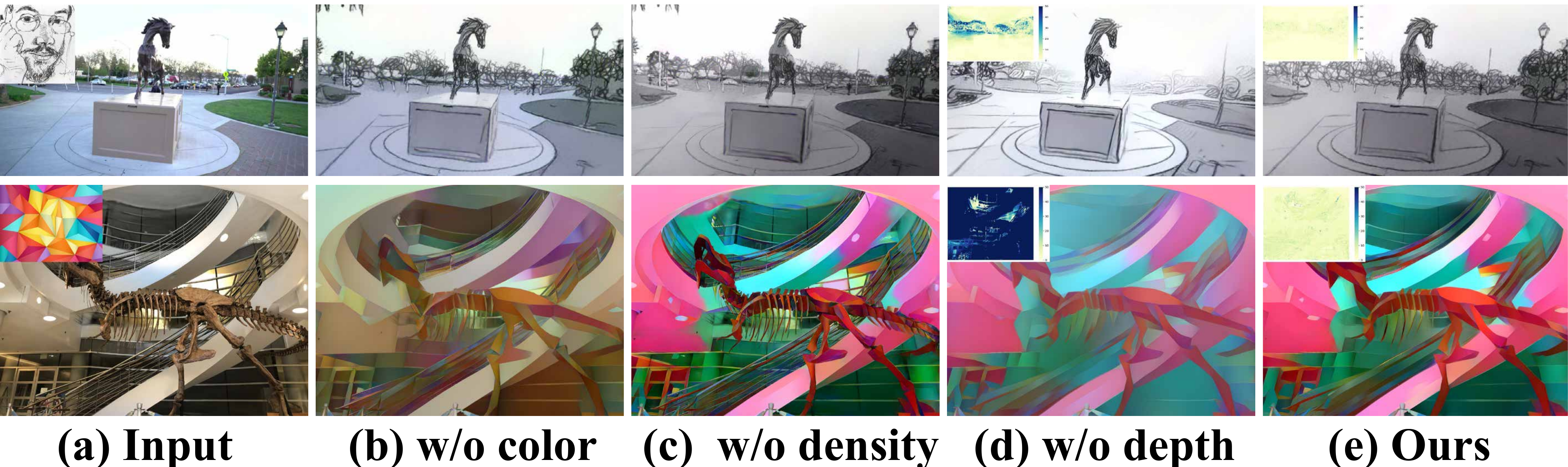}
    \caption{\textbf{Ablation Study about density and color control}. `w/o density' presents the stylized scene without density component fine-tuning in 3DGS. `w/o recolor' displays the stylized results without applying recoloring while `w/o depth' shows the results without incorporating depth loss, leading to the disappearance of original geometry and semantic content.}
    \label{fig:ablation}
    \vspace{-4mm}
\end{figure}

\begin{figure}[ht]
    \centering
    \includegraphics[width=\linewidth]{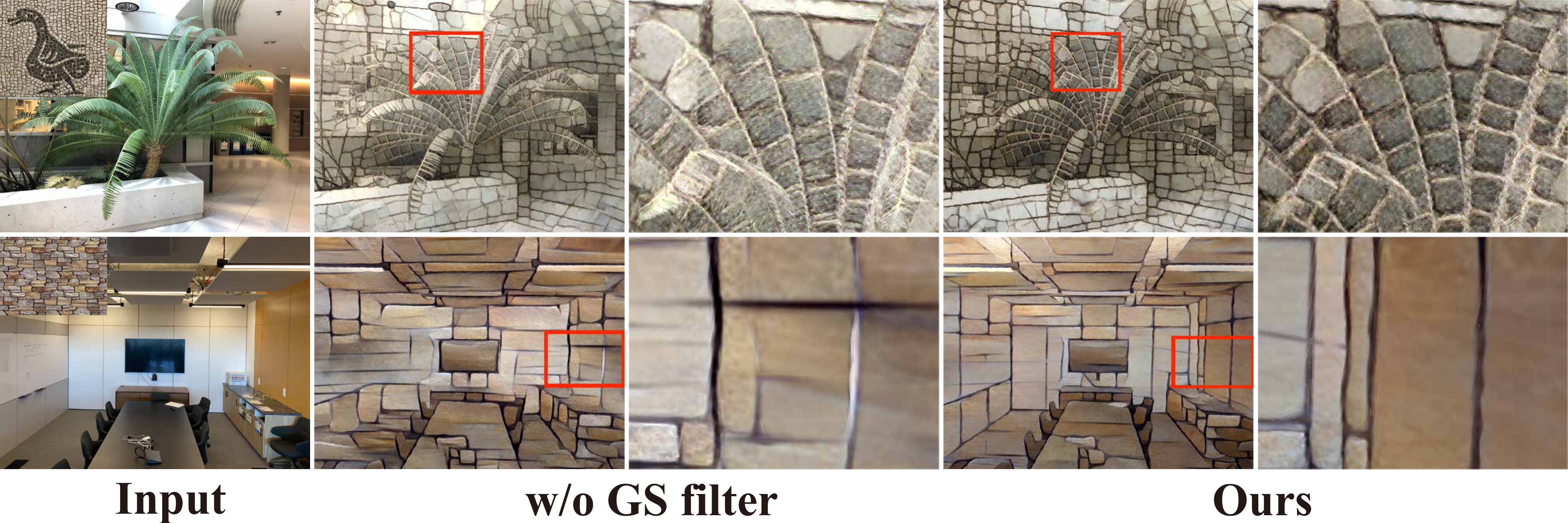}
    \caption{\textbf{Ablation study about 3D Gaussian filter.} The filter effectively helps eliminate artifacts.}
    \label{fig:ablation2}
    \vspace{-4mm}
\end{figure}

\textbf{User Study}
\zdxn{We conducted a comprehensive user study with 52 participants evaluating 40 stylized 3D scenes across multiple datasets. Participants compared our method against baselines using four metrics: visual preference, style fidelity, content preservation, and 3D consistency. Each participant assessed randomized method pairs alongside style references and original scenes, yielding 1000 total votes. Statistical analysis confirms our approach's superiority across all metrics ($p \leq 0.0001$), with results demonstrating significant preference for our method in visual quality and consistency (Fig. \ref{fig:user_study_new}). Study design ensured demographic diversity, unbiased evaluation conditions, and rigorous statistical validation, as detailed in the supplement.}

\subsection{Controllable Stylization Results}
Our method enables users to control stylization in three ways, including color, scale, and spatial region. The corresponding controllable stylization results are shown in Figs.~\ref{fig:color_crtl}, \ref{fig:scale_crtl} and \ref{fig:spatial_crtl}, respectively.
Fig.~\ref{fig:serial_ctrl} shows an example where a sequential series of controls are applied to a scene across different conceptual factors. In addition to the direct stylization result, we sequentially apply different controls from left to right: preserving the color of the original scene, increasing the scale of the pattern style, and adopting spatial control to apply two styles. Such flexibility allows for controllable refinement of key parameters or stylistic attributes, satisfying specific aesthetic preferences or artistic requirements and fostering enhanced artistic expression.

\subsection{Ablation Study}
We conducted ablation studies to validate our design choices. The original scene and input style image are presented in Fig.~\ref{fig:ablation} (a). When the recoloring procedure is not applied, as seen in (b), the stylized scene exhibits inferior color correspondence with the style image.
In Fig.~\ref{fig:ablation} (d)(e), we illustrate the importance of our depth preservation loss in preserving the original scene's geometry. Without applying depth loss, the background of the Horse scene disappears, and the Trex's skeleton becomes blurrier. The heatmap in the top left corner depicts the difference between the rendered depth maps and ground truth depth maps. The notable differences in the heatmap of (d) further emphasize the efficacy of our depth preservation loss.
In Fig.~\ref{fig:ablation} (c), we highlight the importance of fine-tuning both the color and density components in 3DGS, which is crucial for learning intricate style patterns. Neglecting the optimization of the density makes it challenging to capture stroke details and texture patterns in the style image. 
Therefore, a balance is established wherein fine-tuning the density component harmonizes with the depth preservation loss, thereby facilitating the transfer of intricate style details while keeping the integrity of the original geometry. 

Additionally, we investigate how the proposed 3DGS filter affects the reconstructed scenes. Normally, floaters become more noticeable after the style transfer process. As shown in Fig.~\ref{fig:ablation2}, the texture of stones and leaves extend into the air around the objects without the 3DGS filter. In contrast, the results of our method exhibit enhanced clarity, with fewer visible colored floaters.
Tab.~\ref{tab:ablation} shows the quantitative results of the ablation studies, demonstrating the contributions of all components of our method.

\section{Discussion and Conclusion}

\begin{figure}[t]
    \centering
    \includegraphics[width=\linewidth]{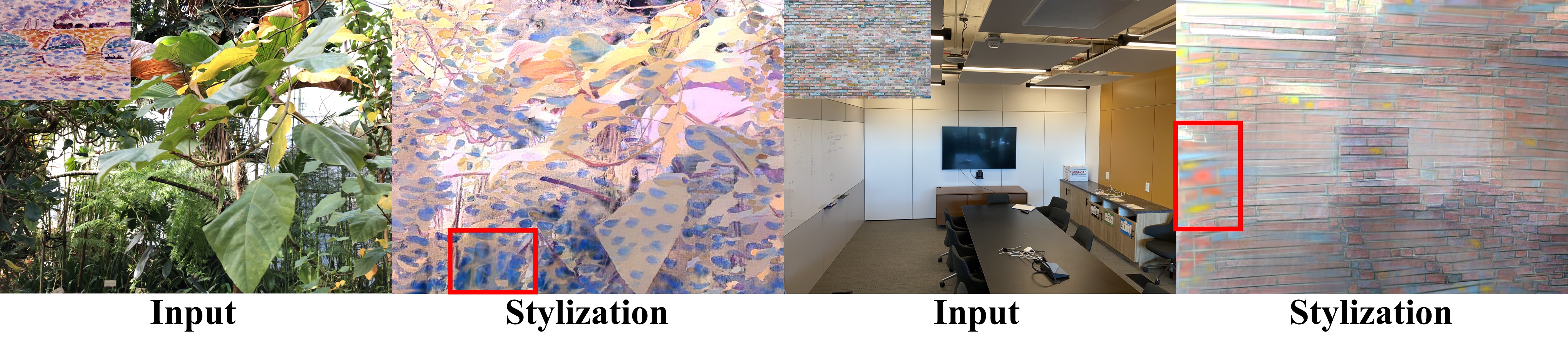}
    \vspace{-6mm}
    \caption{\textbf{Limitation}. The geometric artifacts from the original 3DGS reconstruction may impact the quality of the final stylized scenes.}
    \vspace{-4mm}
    \label{fig:limitation}
\end{figure}

\textbf{Limitations.} Although our method achieves efficient and controllable 3DGS stylization, it still has some limitations. First, the geometric artifacts from the original 3DGS reconstruction may impact the quality of the final stylized scenes, as shown in Fig.~\ref{fig:limitation}. Although our filter-based refinement can eliminate some floaters, it cannot eliminate all geometric artifacts. In the future, we will incorporate some improved 3DGS reconstruction methods that add geometric constraints, such as SuGaR~\cite{Guedon2023SuGaRSG}, to address this issue. In addition, our optimization-based method cannot achieve instant stylization, However, this does not compromise the effectiveness of our method or its practical value, as the optimization process typically takes only about 1 minute. 

In this paper, we introduce \name, the first controllable 3D scene stylization method based on the 3D Gaussian Splatting representation. Following 3DGS reconstruction, our proposed filter-based refinement minimizes the impact of floaters in the reconstruction, which is crucial for achieving the desired stylization effect. Additionally, we propose the adoption of nearest-neighbor feature matching style loss to optimize both geometry and color parameters of 3D Gaussians, enabling the capture of detailed style features and facilitating 3D scene stylization. We further introduce a depth preservation loss for regularization to maintain the overall structure during stylization. Moreover, we design controllable methods for three perceptual factors: color, stylization scale, and spatial regions, providing users with specific and diverse control options. Qualitative and quantitative experiments demonstrate that our method outperforms existing 3D stylization methods in terms of effectiveness and efficiency.

\ifCLASSOPTIONcaptionsoff
  \newpage
\fi

{\small
    \bibliographystyle{IEEEtran}
    \bibliography{main}

\begin{thebibliography}{10}
\providecommand{\url}[1]{#1}
\csname url@samestyle\endcsname
\providecommand{\newblock}{\relax}
\providecommand{\bibinfo}[2]{#2}
\providecommand{\BIBentrySTDinterwordspacing}{\spaceskip=0pt\relax}
\providecommand{\BIBentryALTinterwordstretchfactor}{4}
\providecommand{\BIBentryALTinterwordspacing}{\spaceskip=\fontdimen2\font plus
\BIBentryALTinterwordstretchfactor\fontdimen3\font minus \fontdimen4\font\relax}
\providecommand{\BIBforeignlanguage}[2]{{%
\expandafter\ifx\csname l@#1\endcsname\relax
\typeout{** WARNING: IEEEtran.bst: No hyphenation pattern has been}%
\typeout{** loaded for the language `#1'. Using the pattern for}%
\typeout{** the default language instead.}%
\else
\language=\csname l@#1\endcsname
\fi
#2}}
\providecommand{\BIBdecl}{\relax}
\BIBdecl

\bibitem{Kato_2018_CVPR}
H.~Kato, Y.~Ushiku, and T.~Harada, ``Neural 3d mesh renderer,'' in \emph{Proceedings of the IEEE Conference on Computer Vision and Pattern Recognition (CVPR)}, June 2018.

\bibitem{Michel_2022_CVPR}
O.~Michel, R.~Bar-On, R.~Liu, S.~Benaim, and R.~Hanocka, ``Text2mesh: Text-driven neural stylization for meshes,'' in \emph{Proceedings of the IEEE/CVF Conference on Computer Vision and Pattern Recognition (CVPR)}, June 2022, pp. 13\,492--13\,502.

\bibitem{yin20213dstylenet}
K.~Yin, J.~Gao, M.~Shugrina, S.~Khamis, and S.~Fidler, ``3dstylenet: Creating 3d shapes with geometric and texture style variations,'' in \emph{Proceedings of the IEEE/CVF International Conference on Computer Vision}, 2021, pp. 12\,456--12\,465.

\bibitem{guo2021volumetric}
J.~Guo, M.~Li, Z.~Zong, Y.~Liu, J.~He, Y.~Guo, and L.-Q. Yan, ``Volumetric appearance stylization with stylizing kernel prediction network.'' \emph{ACM Trans. Graph.}, vol.~40, no.~4, pp. 162--1, 2021.

\bibitem{klehm2014property}
O.~Klehm, I.~Ihrke, H.-P. Seidel, and E.~Eisemann, ``Property and lighting manipulations for static volume stylization using a painting metaphor,'' \emph{IEEE Transactions on Visualization and Computer Graphics}, vol.~20, no.~7, pp. 983--995, 2014.

\bibitem{cao2020psnet}
X.~Cao, W.~Wang, K.~Nagao, and R.~Nakamura, ``Psnet: A style transfer network for point cloud stylization on geometry and color,'' in \emph{Proceedings of the IEEE/CVF Winter Conference on Applications of Computer vision}, 2020, pp. 3337--3345.

\bibitem{huang2021learning}
H.-P. Huang, H.-Y. Tseng, S.~Saini, M.~Singh, and M.-H. Yang, ``Learning to stylize novel views,'' in \emph{Proceedings of the IEEE/CVF International Conference on Computer Vision}, 2021, pp. 13\,869--13\,878.

\bibitem{bae2023point}
E.~Bae, J.~Kim, and S.~Lee, ``Point cloud-based free viewpoint artistic style transfer,'' in \emph{2023 IEEE International Conference on Multimedia and Expo Workshops (ICMEW)}.\hskip 1em plus 0.5em minus 0.4em\relax IEEE, 2023, pp. 302--307.

\bibitem{chiang2022stylizing}
P.-Z. Chiang, M.-S. Tsai, H.-Y. Tseng, W.-S. Lai, and W.-C. Chiu, ``Stylizing 3d scene via implicit representation and hypernetwork,'' in \emph{Proceedings of the IEEE/CVF Winter Conference on Applications of Computer Vision}, 2022, pp. 1475--1484.

\bibitem{nguyen2022snerf}
T.~Nguyen-Phuoc, F.~Liu, and L.~Xiao, ``Snerf: stylized neural implicit representations for 3d scenes,'' \emph{ACM Transactions on Graphics (TOG)}, vol.~41, no.~4, pp. 1--11, 2022.

\bibitem{fan2022unified}
Z.~Fan, Y.~Jiang, P.~Wang, X.~Gong, D.~Xu, and Z.~Wang, ``Unified implicit neural stylization,'' in \emph{European Conference on Computer Vision}.\hskip 1em plus 0.5em minus 0.4em\relax Springer, 2022, pp. 636--654.

\bibitem{huang2022stylizednerf}
Y.-H. Huang, Y.~He, Y.-J. Yuan, Y.-K. Lai, and L.~Gao, ``Stylizednerf: consistent 3d scene stylization as stylized nerf via 2d-3d mutual learning,'' in \emph{Proceedings of the IEEE/CVF Conference on Computer Vision and Pattern Recognition}, 2022, pp. 18\,342--18\,352.

\bibitem{zhang2022arf}
K.~Zhang, N.~Kolkin, S.~Bi, F.~Luan, Z.~Xu, E.~Shechtman, and N.~Snavely, ``Arf: Artistic radiance fields,'' in \emph{ECCV}, 2022.

\bibitem{wang2023nerf}
C.~Wang, R.~Jiang, M.~Chai, M.~He, D.~Chen, and J.~Liao, ``Nerf-art: Text-driven neural radiance fields stylization,'' \emph{IEEE Transactions on Visualization and Computer Graphics}, 2023.

\bibitem{pang2023locally}
H.-W. Pang, B.-S. Hua, and S.-K. Yeung, ``Locally stylized neural radiance fields,'' in \emph{2023 IEEE/CVF International Conference on Computer Vision (ICCV)}.\hskip 1em plus 0.5em minus 0.4em\relax IEEE Computer Society, 2023, pp. 307--316.

\bibitem{zhang2023transforming}
Z.~Zhang, Y.~Liu, C.~Han, Y.~Pan, T.~Guo, and T.~Yao, ``Transforming radiance field with lipschitz network for photorealistic 3d scene stylization,'' in \emph{Proceedings of the IEEE/CVF Conference on Computer Vision and Pattern Recognition}, 2023, pp. 20\,712--20\,721.

\bibitem{kerbl3Dgaussians}
\BIBentryALTinterwordspacing
B.~Kerbl, G.~Kopanas, T.~Leimk{\"u}hler, and G.~Drettakis, ``3d gaussian splatting for real-time radiance field rendering,'' \emph{ACM Transactions on Graphics}, vol.~42, no.~4, July 2023. [Online]. Available: \url{https://repo-sam.inria.fr/fungraph/3d-gaussian-splatting/}
\BIBentrySTDinterwordspacing

\bibitem{gaussian_grouping}
M.~Ye, M.~Danelljan, F.~Yu, and L.~Ke, ``Gaussian grouping: Segment and edit anything in 3d scenes,'' \emph{arXiv preprint arXiv:2312.00732}, 2023.

\bibitem{GaussianEditor}
J.~Fang, J.~Wang, X.~Zhang, L.~Xie, and Q.~Tian, ``Gaussianeditor: Editing 3d gaussians delicately with text instructions,'' \emph{arXiv preprint arXiv:2311.16037}, 2023.

\bibitem{chen2023gaussianeditor}
Y.~Chen, Z.~Chen, C.~Zhang, F.~Wang, X.~Yang, Y.~Wang, Z.~Cai, L.~Yang, H.~Liu, and G.~Lin, ``Gaussianeditor: Swift and controllable 3d editing with gaussian splatting,'' 2023.

\bibitem{tang2023dreamgaussian}
J.~Tang, J.~Ren, H.~Zhou, Z.~Liu, and G.~Zeng, ``Dreamgaussian: Generative gaussian splatting for efficient 3d content creation,'' \emph{arXiv preprint arXiv:2309.16653}, 2023.

\bibitem{liu2024stylegaussian}
K.~Liu, F.~Zhan, M.~Xu, C.~Theobalt, L.~Shao, and S.~Lu, ``Stylegaussian: Instant 3d style transfer with gaussian splatting,'' \emph{arXiv preprint arXiv:2403.07807}, 2024.

\bibitem{saroha2024gaussian}
A.~Saroha, M.~Gladkova, C.~Curreli, T.~Yenamandra, and D.~Cremers, ``Gaussian splatting in style,'' \emph{arXiv preprint arXiv:2403.08498}, 2024.

\bibitem{gatys2017controlling}
L.~A. Gatys, A.~S. Ecker, M.~Bethge, A.~Hertzmann, and E.~Shechtman, ``Controlling perceptual factors in neural style transfer,'' in \emph{Proceedings of the IEEE conference on computer vision and pattern recognition}, 2017, pp. 3985--3993.

\bibitem{jing2018stroke}
Y.~Jing, Y.~Liu, Y.~Yang, Z.~Feng, Y.~Yu, D.~Tao, and M.~Song, ``Stroke controllable fast style transfer with adaptive receptive fields,'' in \emph{Proceedings of the European Conference on Computer Vision (ECCV)}, 2018, pp. 238--254.

\bibitem{castillo2017zorn}
C.~Castillo, S.~De, X.~Han, B.~Singh, A.~K. Yadav, and T.~Goldstein, ``Son of zorn's lemma: Targeted style transfer using instance-aware semantic segmentation,'' in \emph{2017 IEEE International Conference on Acoustics, Speech and Signal Processing (ICASSP)}.\hskip 1em plus 0.5em minus 0.4em\relax IEEE, 2017, pp. 1348--1352.

\bibitem{li2023arf}
W.~Li, T.~Wu, F.~Zhong, and C.~Oztireli, ``Arf-plus: Controlling perceptual factors in artistic radiance fields for 3d scene stylization,'' \emph{arXiv preprint arXiv:2308.12452}, 2023.

\bibitem{gatys2015neural}
L.~A. Gatys, A.~S. Ecker, and M.~Bethge, ``A neural algorithm of artistic style,'' \emph{arXiv preprint arXiv:1508.06576}, 2015.

\bibitem{gatys2016image}
------, ``Image style transfer using convolutional neural networks,'' in \emph{Proceedings of the IEEE conference on computer vision and pattern recognition}, 2016, pp. 2414--2423.

\bibitem{simonyan2014very}
K.~Simonyan and A.~Zisserman, ``Very deep convolutional networks for large-scale image recognition,'' \emph{arXiv preprint arXiv:1409.1556}, 2014.

\bibitem{risser2017stable}
E.~Risser, P.~Wilmot, and C.~Barnes, ``Stable and controllable neural texture synthesis and style transfer using histogram losses,'' \emph{arXiv preprint arXiv:1701.08893}, 2017.

\bibitem{gu2018arbitrary}
S.~Gu, C.~Chen, J.~Liao, and L.~Yuan, ``Arbitrary style transfer with deep feature reshuffle,'' in \emph{Proceedings of the IEEE Conference on Computer Vision and Pattern Recognition}, 2018, pp. 8222--8231.

\bibitem{kolkin2019style}
N.~Kolkin, J.~Salavon, and G.~Shakhnarovich, ``Style transfer by relaxed optimal transport and self-similarity,'' in \emph{Proceedings of the IEEE/CVF Conference on Computer Vision and Pattern Recognition}, 2019, pp. 10\,051--10\,060.

\bibitem{liao2017visual}
J.~Liao, Y.~Yao, L.~Yuan, G.~Hua, and S.~B. Kang, ``Visual attribute transfer through deep image analogy,'' \emph{arXiv preprint arXiv:1705.01088}, 2017.

\bibitem{an2021artflow}
J.~An, S.~Huang, Y.~Song, D.~Dou, W.~Liu, and J.~Luo, ``Artflow: Unbiased image style transfer via reversible neural flows,'' in \emph{Proceedings of the IEEE/CVF Conference on Computer Vision and Pattern Recognition}, 2021, pp. 862--871.

\bibitem{huang2017arbitrary}
X.~Huang and S.~Belongie, ``Arbitrary style transfer in real-time with adaptive instance normalization,'' in \emph{Proceedings of the IEEE international conference on computer vision}, 2017, pp. 1501--1510.

\bibitem{park2019arbitrary}
D.~Y. Park and K.~H. Lee, ``Arbitrary style transfer with style-attentional networks,'' in \emph{proceedings of the IEEE/CVF conference on computer vision and pattern recognition}, 2019, pp. 5880--5888.

\bibitem{barnes2009patchmatch}
C.~Barnes, E.~Shechtman, A.~Finkelstein, and D.~B. Goldman, ``Patchmatch: A randomized correspondence algorithm for structural image editing,'' \emph{ACM Trans. Graph.}, vol.~28, no.~3, p.~24, 2009.

\bibitem{chen2016fast}
T.~Q. Chen and M.~Schmidt, ``Fast patch-based style transfer of arbitrary style,'' \emph{arXiv preprint arXiv:1612.04337}, 2016.

\bibitem{Zhang_2023_inst}
Y.~Zhang, N.~Huang, F.~Tang, H.~Huang, C.~Ma, W.~Dong, and C.~Xu, ``Inversion-based style transfer with diffusion models,'' in \emph{Proceedings of the IEEE/CVF Conference on Computer Vision and Pattern Recognition (CVPR)}, June 2023, pp. 10\,146--10\,156.

\bibitem{he2024multi}
Y.~He, L.~Chen, Y.-J. Yuan, S.-Y. Chen, and L.~Gao, ``Multi-level patch transformer for style transfer with single reference image,'' in \emph{International Conference on Computational Visual Media}.\hskip 1em plus 0.5em minus 0.4em\relax Springer, 2024, pp. 221--239.

\bibitem{zhu2017unpaired}
J.-Y. Zhu, T.~Park, P.~Isola, and A.~A. Efros, ``Unpaired image-to-image translation using cycle-consistent adversarial networks,'' in \emph{Proceedings of the IEEE international conference on computer vision}, 2017, pp. 2223--2232.

\bibitem{vaswani2017attention}
A.~Vaswani, N.~Shazeer, N.~Parmar, J.~Uszkoreit, L.~Jones, A.~N. Gomez, {\L}.~Kaiser, and I.~Polosukhin, ``Attention is all you need,'' in \emph{Advances in neural information processing systems}, 2017, pp. 5998--6008.

\bibitem{rombach2022high}
R.~Rombach, A.~Blattmann, D.~Lorenz, P.~Esser, and B.~Ommer, ``High-resolution image synthesis with latent diffusion models,'' in \emph{Proceedings of the IEEE/CVF conference on computer vision and pattern recognition}, 2022, pp. 10\,684--10\,695.

\bibitem{cen2023saga}
J.~Cen, J.~Fang, C.~Yang, L.~Xie, X.~Zhang, W.~Shen, and Q.~Tian, ``Segment any 3d gaussians,'' \emph{arXiv preprint arXiv:2312.00860}, 2023.

\bibitem{gao2024mesh}
L.~Gao, J.~Yang, B.-T. Zhang, J.-M. Sun, Y.-J. Yuan, H.~Fu, and Y.-K. Lai, ``Mesh-based gaussian splatting for real-time large-scale deformation,'' \emph{arXiv preprint arXiv:2402.04796}, 2024.

\bibitem{wu2024deferredgs}
T.~Wu, J.-M. Sun, Y.-K. Lai, Y.~Ma, L.~Kobbelt, and L.~Gao, ``Deferredgs: Decoupled and editable gaussian splatting with deferred shading,'' \emph{arXiv preprint arXiv:2404.09412}, 2024.

\bibitem{wu2024recent}
T.~Wu, Y.-J. Yuan, L.-X. Zhang, J.~Yang, Y.-P. Cao, L.-Q. Yan, and L.~Gao, ``Recent advances in 3d gaussian splatting,'' \emph{arXiv preprint arXiv:2403.11134}, 2024.

\bibitem{chen2024survey}
G.~Chen and W.~Wang, ``A survey on 3d gaussian splatting,'' \emph{arXiv preprint arXiv:2401.03890}, 2024.

\bibitem{mildenhall2021nerf}
B.~Mildenhall, P.~P. Srinivasan, M.~Tancik, J.~T. Barron, R.~Ramamoorthi, and R.~Ng, ``Nerf: Representing scenes as neural radiance fields for view synthesis,'' \emph{Communications of the ACM}, vol.~65, no.~1, pp. 99--106, 2021.

\bibitem{xu2023desrf}
S.~Xu, L.~Li, L.~Shen, and Z.~Lian, ``Desrf: Deformable stylized radiance field,'' in \emph{Proceedings of the IEEE/CVF Conference on Computer Vision and Pattern Recognition}, 2023, pp. 709--718.

\bibitem{kumar2023s2rf}
M.~Kumar, N.~Panse, and D.~Lahiri, ``S2rf: Semantically stylized radiance fields,'' in \emph{Proceedings of the IEEE/CVF International Conference on Computer Vision}, 2023, pp. 2952--2957.

\bibitem{Zhang_2023_CVPR}
Y.~Zhang, Z.~He, J.~Xing, X.~Yao, and J.~Jia, ``Ref-npr: Reference-based non-photorealistic radiance fields for controllable scene stylization,'' in \emph{Proceedings of the IEEE/CVF Conference on Computer Vision and Pattern Recognition (CVPR)}, June 2023, pp. 4242--4251.

\bibitem{karras2019style}
T.~Karras, S.~Laine, and T.~Aila, ``A style-based generator architecture for generative adversarial networks,'' in \emph{Proceedings of the IEEE/CVF conference on computer vision and pattern recognition}, 2019, pp. 4401--4410.

\bibitem{kolkin2022neural}
N.~Kolkin, M.~Kucera, S.~Paris, D.~Sykora, E.~Shechtman, and G.~Shakhnarovich, ``Neural neighbor style transfer,'' \emph{arXiv e-prints}, pp. arXiv--2203, 2022.

\bibitem{kirillov2023segany}
A.~Kirillov, E.~Mintun, N.~Ravi, H.~Mao, C.~Rolland, L.~Gustafson, T.~Xiao, S.~Whitehead, A.~C. Berg, W.-Y. Lo, P.~Doll{\'a}r, and R.~Girshick, ``Segment anything,'' \emph{arXiv:2304.02643}, 2023.

\bibitem{langsam_github}
``Language segment-anything,'' 2023, https://github.com/luca-medeiros/lang-segment-anything.

\bibitem{mildenhall2019llff}
B.~Mildenhall, P.~P. Srinivasan, R.~Ortiz-Cayon, N.~K. Kalantari, R.~Ramamoorthi, R.~Ng, and A.~Kar, ``Local light field fusion: Practical view synthesis with prescriptive sampling guidelines,'' \emph{ACM Transactions on Graphics (TOG)}, 2019.

\bibitem{knapitsch2017tanks}
A.~Knapitsch, J.~Park, Q.-Y. Zhou, and V.~Koltun, ``Tanks and temples: Benchmarking large-scale scene reconstruction,'' \emph{ACM Transactions on Graphics (ToG)}, vol.~36, no.~4, pp. 1--13, 2017.

\bibitem{barron2022mipnerf360}
J.~T. Barron, B.~Mildenhall, D.~Verbin, P.~P. Srinivasan, and P.~Hedman, ``Mip-nerf 360: Unbounded anti-aliased neural radiance fields,'' \emph{CVPR}, 2022.

\bibitem{DeepBlending2018}
P.~Hedman, J.~Philip, T.~Price, J.-M. Frahm, G.~Drettakis, and G.~Brostow, ``Deep blending for free-viewpoint image-based rendering,'' in \emph{ACM Transactions on Graphics (Proc. SIGGRAPH Asia)}, vol.~37, no.~6, 2018, pp. 257:1--257:15.

\bibitem{instructnerf2023}
A.~Haque, M.~Tancik, A.~Efros, A.~Holynski, and A.~Kanazawa, ``Instruct-nerf2nerf: Editing 3d scenes with instructions,'' in \emph{Proceedings of the IEEE/CVF International Conference on Computer Vision}, 2023.

\bibitem{artgan2018}
\BIBentryALTinterwordspacing
W.~R. Tan, C.~S. Chan, H.~Aguirre, and K.~Tanaka, ``Improved artgan for conditional synthesis of natural image and artwork,'' \emph{IEEE Transactions on Image Processing}, vol.~28, no.~1, pp. 394--409, 2019. [Online]. Available: \url{https://doi.org/10.1109/TIP.2018.2866698}
\BIBentrySTDinterwordspacing

\bibitem{liu2023stylerf}
K.~Liu, F.~Zhan, Y.~Chen, J.~Zhang, Y.~Yu, A.~El~Saddik, S.~Lu, and E.~P. Xing, ``Stylerf: Zero-shot 3d style transfer of neural radiance fields,'' in \emph{Proceedings of the IEEE/CVF Conference on Computer Vision and Pattern Recognition}, 2023, pp. 8338--8348.

\bibitem{wright2022artfid}
M.~Wright and B.~Ommer, ``Artfid: Quantitative evaluation of neural style transfer,'' in \emph{DAGM German Conference on Pattern Recognition}.\hskip 1em plus 0.5em minus 0.4em\relax Springer, 2022, pp. 560--576.

\bibitem{huang2023quantart}
S.~Huang, J.~An, D.~Wei, J.~Luo, and H.~Pfister, ``Quantart: Quantizing image style transfer towards high visual fidelity,'' in \emph{Proceedings of the IEEE/CVF Conference on Computer Vision and Pattern Recognition}, 2023, pp. 5947--5956.

\bibitem{ssim}
Z.~Wang, A.~Bovik, H.~Sheikh, and E.~Simoncelli, ``Image quality assessment: from error visibility to structural similarity,'' \emph{IEEE Transactions on Image Processing}, vol.~13, no.~4, pp. 600--612, 2004.

\bibitem{ding2020iqa}
\BIBentryALTinterwordspacing
K.~Ding, K.~Ma, S.~Wang, and E.~P. Simoncelli, ``Image quality assessment: Unifying structure and texture similarity,'' \emph{CoRR}, vol. abs/2004.07728, 2020. [Online]. Available: \url{https://arxiv.org/abs/2004.07728}
\BIBentrySTDinterwordspacing

\bibitem{hong2021domain}
K.~Hong, S.~Jeon, H.~Yang, J.~Fu, and H.~Byun, ``Domain-aware universal style transfer,'' in \emph{Proceedings of the IEEE/CVF International Conference on Computer Vision}, 2021, pp. 14\,609--14\,617.

\bibitem{yang2022industrial}
J.~Yang, F.~Guo, S.~Chen, J.~Li, and J.~Yang, ``Industrial style transfer with large-scale geometric warping and content preservation,'' in \emph{Proceedings of the IEEE/CVF Conference on Computer Vision and Pattern Recognition}, 2022, pp. 7834--7843.

\bibitem{kim2021deep}
S.~Kim, S.~Kim, and S.~Kim, ``Deep translation prior: Test-time training for photorealistic style transfer,'' \emph{arXiv preprint arXiv:2112.06150}, 2021.

\bibitem{dist_cite}
Y.~Yu, D.~Li, B.~Li, and N.~Li, ``Multi-style image generation based on semantic image,'' \emph{The Visual Computer}, vol.~40, pp. 1--16, 08 2023.

\bibitem{dhiman2023corf}
A.~Dhiman, R.~Srinath, S.~Sarkar, L.~R. Boregowda, and R.~V. Babu, ``Corf: Colorizing radiance fields using knowledge distillation,'' \emph{arXiv preprint arXiv:2309.07668}, 2023.

\bibitem{teed2020raft}
Z.~Teed and J.~Deng, ``Raft: Recurrent all-pairs field transforms for optical flow,'' in \emph{European conference on computer vision}.\hskip 1em plus 0.5em minus 0.4em\relax Springer, 2020, pp. 402--419.

\bibitem{niklaus2020softmax}
S.~Niklaus and F.~Liu, ``Softmax splatting for video frame interpolation,'' in \emph{Proceedings of the IEEE/CVF Conference on Computer Vision and Pattern Recognition}, 2020, pp. 5437--5446.

\bibitem{zhang2018unreasonable}
R.~Zhang, P.~Isola, A.~A. Efros, E.~Shechtman, and O.~Wang, ``The unreasonable effectiveness of deep features as a perceptual metric,'' in \emph{Proceedings of the IEEE conference on computer vision and pattern recognition}, 2018, pp. 586--595.

\bibitem{Guedon2023SuGaRSG}
\BIBentryALTinterwordspacing
A.~Gu'edon and V.~Lepetit, ``Sugar: Surface-aligned gaussian splatting for efficient 3d mesh reconstruction and high-quality mesh rendering,'' \emph{ArXiv}, vol. abs/2311.12775, 2023. [Online]. Available: \url{https://api.semanticscholar.org/CorpusID:265308825}
\BIBentrySTDinterwordspacing

\bibitem{kingma2014adam}
D.~P. Kingma and J.~Ba, ``Adam: A method for stochastic optimization,'' \emph{arXiv preprint arXiv:1412.6980}, 2014.

\bibitem{huang20242d}
B.~Huang, Z.~Yu, A.~Chen, A.~Geiger, and S.~Gao, ``2d gaussian splatting for geometrically accurate radiance fields,'' \emph{arXiv preprint arXiv:2403.17888}, 2024.

\end{thebibliography}
}

\vspace{-12mm}

\begin{IEEEbiography}
	[{\includegraphics[width=1in,height=1.25in,clip,keepaspectratio]{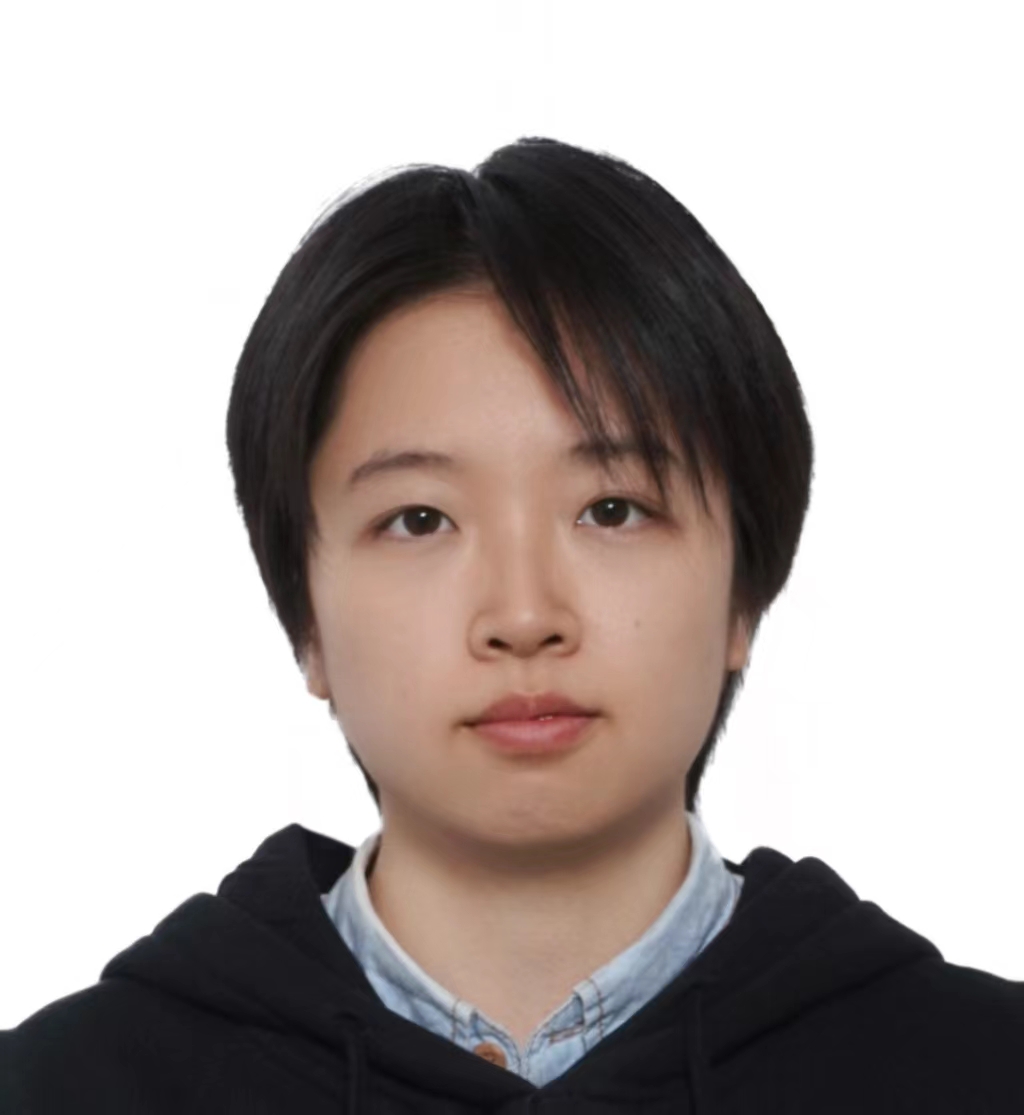}}]
	{Dingxi Zhang} received the bachelor’s degree in Computer Science and Technology from University of Chinese Academy of Sciences in 2024. She is currently pursuing a Master's degree at Department of Computer Science, ETH Zurich. Her reserach interests include 3D vision and computer graphics. 
\end{IEEEbiography}

\vspace{-12mm}

\begin{IEEEbiography}
	[{\includegraphics[width=1in,height=1.25in,clip,keepaspectratio]{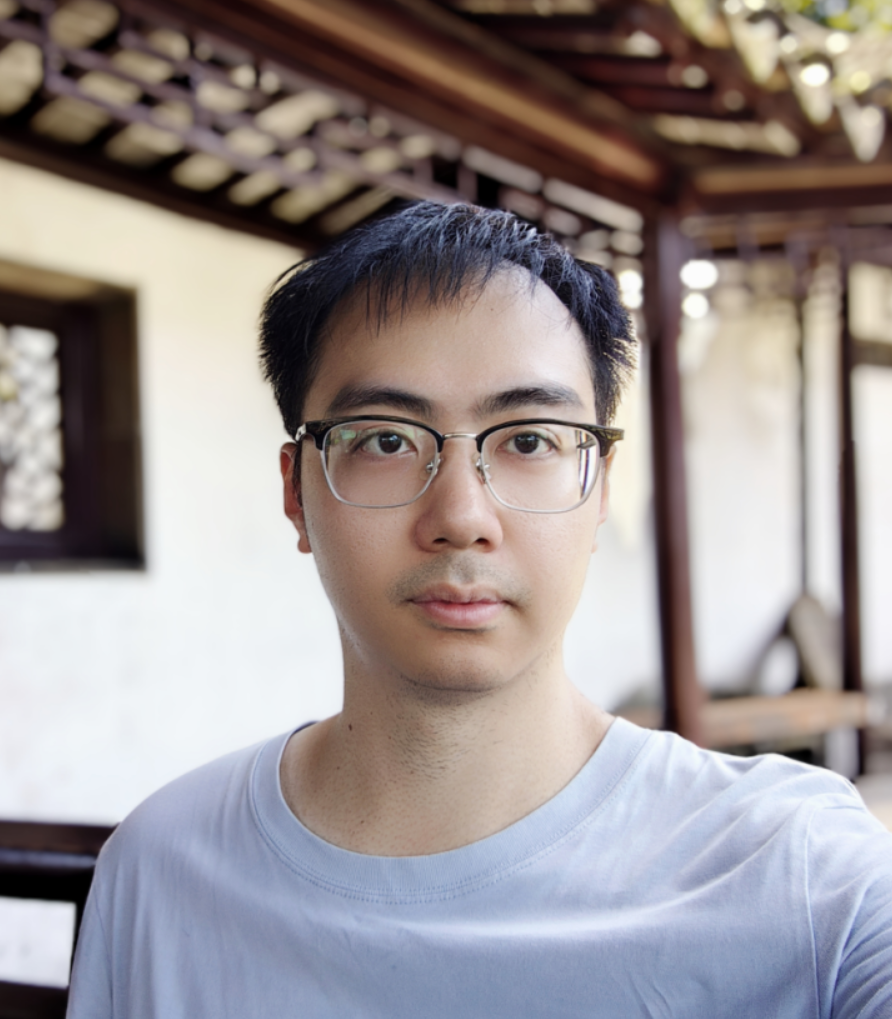}}]
	{Yu-Jie Yuan} received the Ph.D. degree from the Institute of Computing Technology, Chinese Academy of Sciences. His research interests include 3D learning and Multi-Modal Large Language Model.
\end{IEEEbiography}

\vspace{-12mm}

\begin{IEEEbiography}
	[{\includegraphics[width=1in,height=1.25in,clip,keepaspectratio]{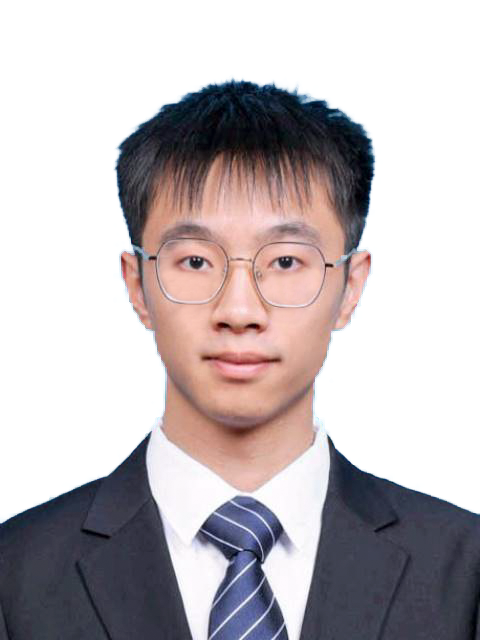}}]
	{Zhuoxun Chen} received a bachelor's degree in Computer Science and Technology from the University of Chinese Academy of Sciences in 2024. He is currently pursuing a Master's degree at the Institute of Computing Technology, Chinese Academy of Sciences. His research interests include neural rendering and robotics.
\end{IEEEbiography}

\vspace{-12mm}

\begin{IEEEbiography}
	[{\includegraphics[width=1in,height=1.25in,clip,keepaspectratio]{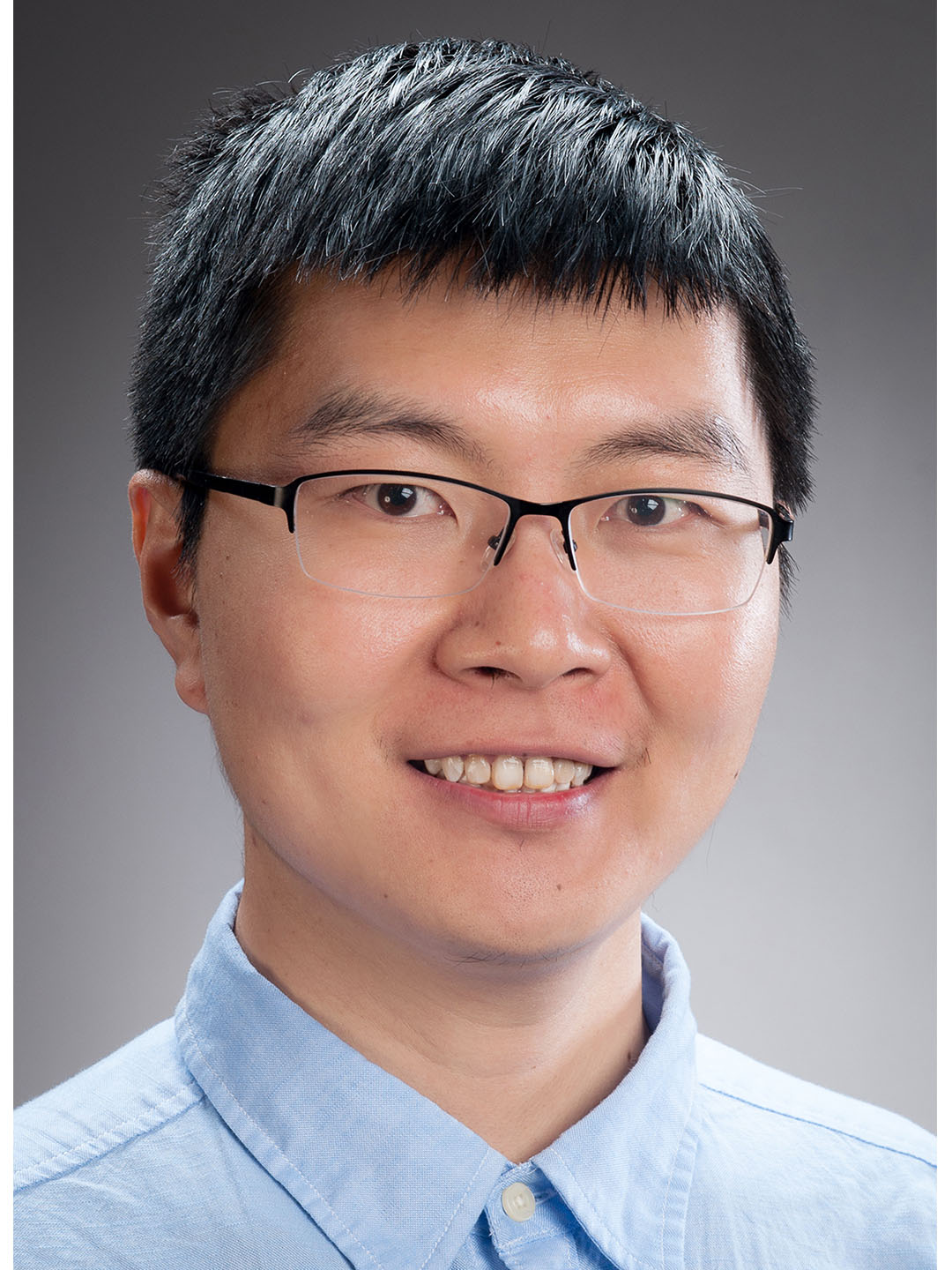}}]
	{Fang-Lue Zhang} is currently a senior lecturer with Victoria University of Wellington, New Zealand. He received the Doctoral degree from Tsinghua University in 2015. His research interests include image and video editing, computer vision, and computer graphics. He received Victoria Early-Career Research Excellence Award in 2019 and Fast-Start Marsden Grant from New Zealand Royal Society in 2020. He is on the editorial board of Computer \& Graphics. He is a committee member of IEEE Central New Zealand Section.
\end{IEEEbiography}

\vspace{-12mm}

\begin{IEEEbiography}
	[{\includegraphics[width=1in,height=1.25in,clip,keepaspectratio]{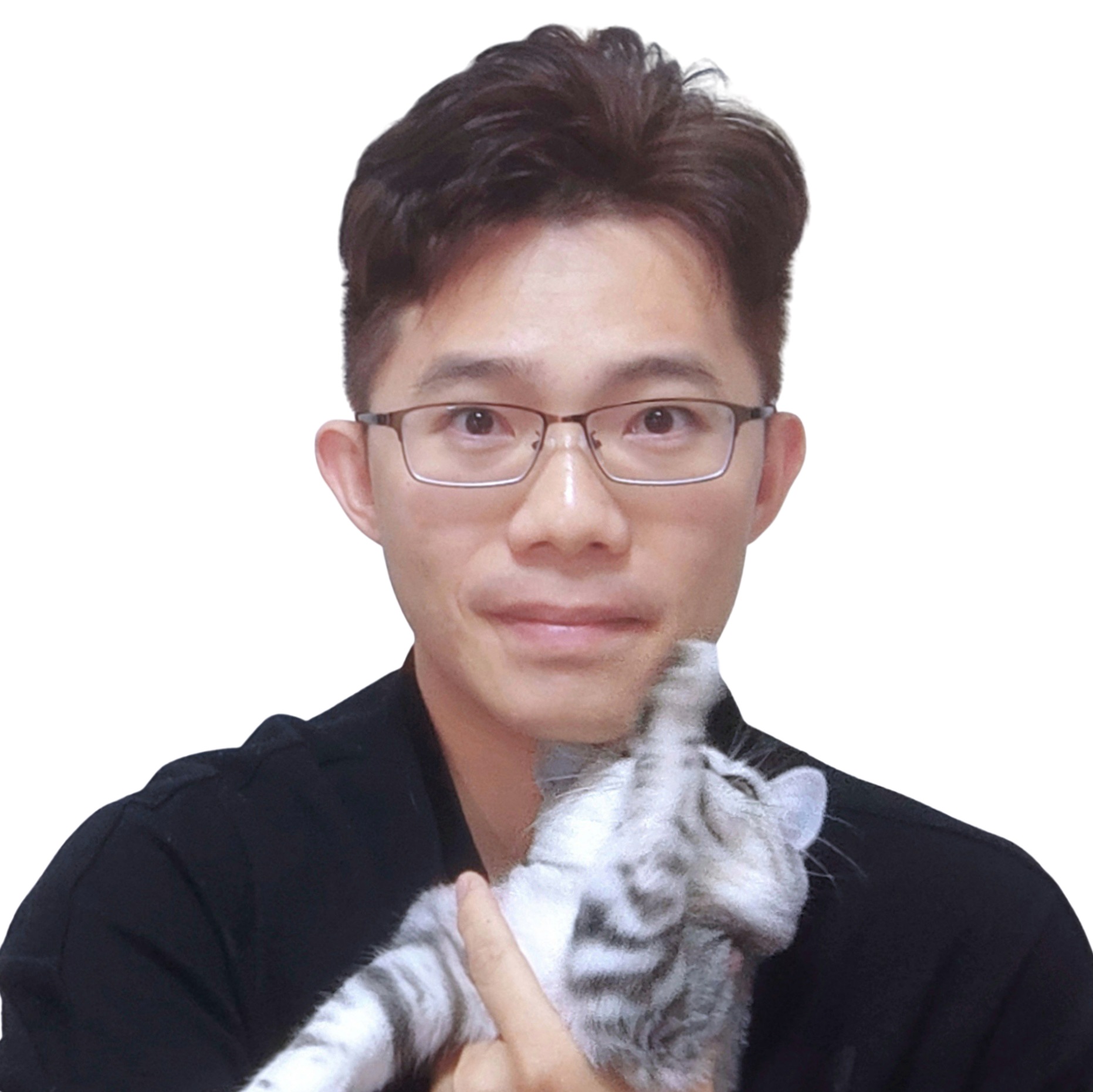}}]
	{Zhenliang He} is currently an Assistant Professor in the Visual Information Processing and Learning research group at the Institute of Computing Technology (ICT), Chinese Academy of Sciences (CAS). He received the Ph.D. from ICT, CAS in 2021 and received the bachelor's degree from Beijing University of Posts and Telecommunications in 2015. His current research focuses on generative models and representation learning.
\end{IEEEbiography}

\vspace{-12mm}

\begin{IEEEbiography}
	[{\includegraphics[width=1in,height=1.25in,clip,keepaspectratio]{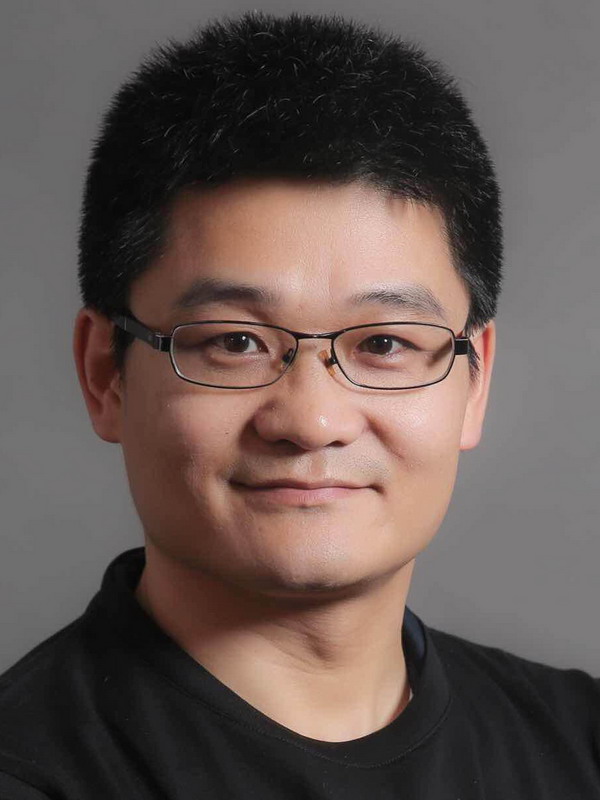}}]
	{Shiguang Shan} received M.S. degree in computer science from the Harbin Institute of Technology, Harbin, China, in 1999, and Ph.D. degree in computer science from the Institute of Computing Technology (ICT), Chinese Academy of Sciences (CAS), Beijing, China, in 2004. He joined ICT, CAS in 2002 and became a full Professor in 2010. He is now the director of the Key Lab of Intelligent Information Processing of CAS.
\end{IEEEbiography}

\vspace{-12mm}

\begin{IEEEbiography}
	[{\includegraphics[width=1in,height=1.25in,clip,keepaspectratio]{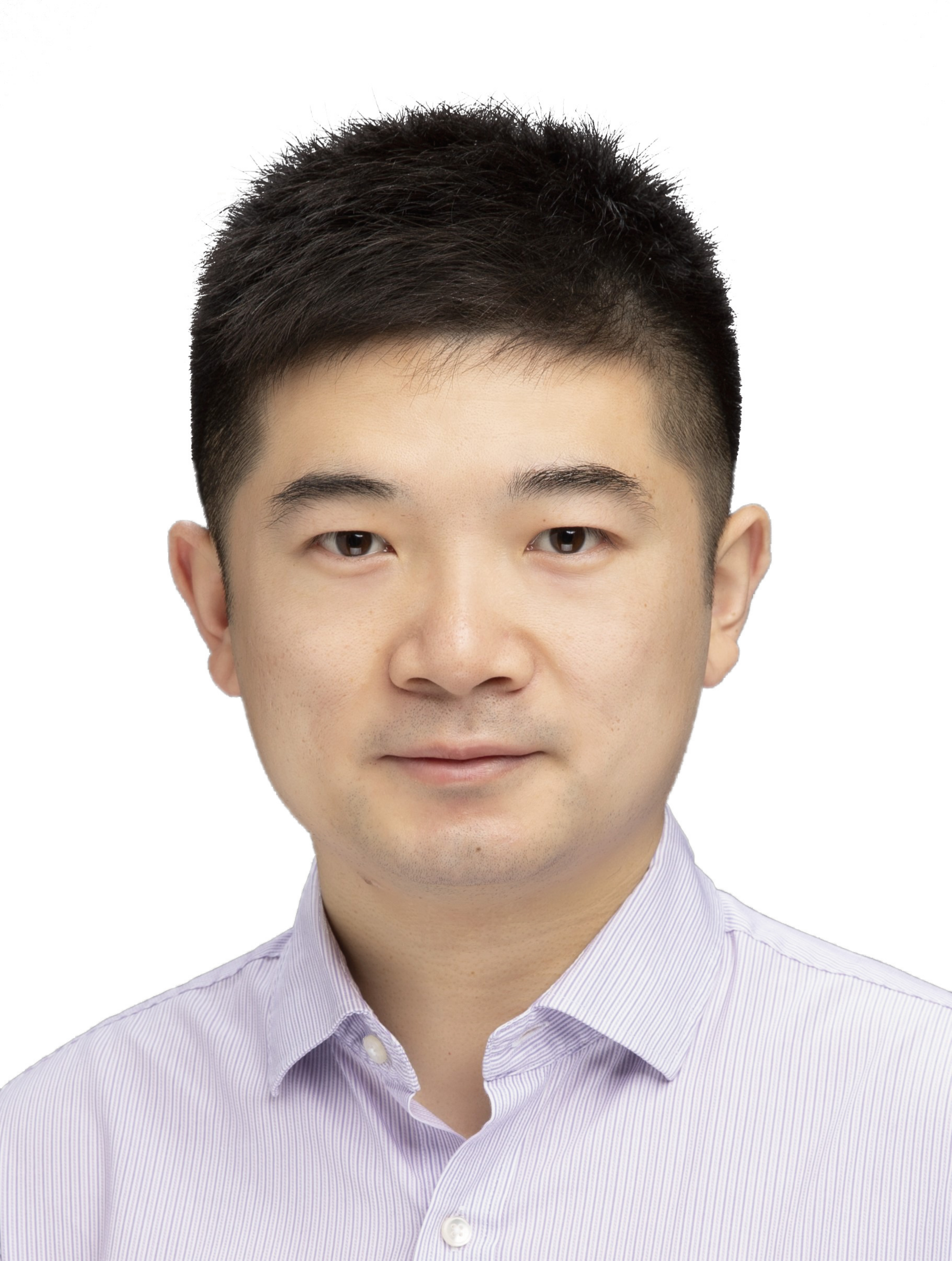}}]
	{Lin Gao} received the PhD degree from Tsinghua University. He is currently a Professor at the Institute of Computing Technology, Chinese Academy of Sciences and the University of Chinese Academy of Sciences. He has been awarded the Newton Advanced Fellowship from Royal Society and Asia Graphics Association Young Researcher Award. His research interests include computer graphics and geometric processing. 
\end{IEEEbiography}

\clearpage
\begin{center}
    \section*{\Large Supplementary Material for \name: Controllable Stylization for 3D Gaussian Splatting}
\end{center}
\setcounter{section}{0} 
\setcounter{subsection}{0} 
\setcounter{figure}{0} 
\setcounter{table}{0}

\section{Overview}
This supplementary document accompanies our main paper and provides implementation details along with additional stylization results. It is divided into four sections: implementation details, more stylization results, additional validation experiments, and a discussion about our future work.

\begin{itemize}
    \item Section~\ref{sec:detail} provides the implementation details, including training settings, more details about the color transfer, spatial control, and user study.
    \item Section~\ref{sec:results} provides additional stylization results, including more controllable stylization results.
    \item Section~\ref{sec:supp_exp} provides additional validation experiments, including more comparisons with baseline methods on LLFF and T\&T datasets, and an additional ablation study.
    \item Section~\ref{sec:future} provides a discussion of our future work.
\end{itemize}

\section{Implementation Details}
\label{sec:detail}
\subsection{Training Settings}
We implement \name based on the 3DGS representation \cite{kerbl3Dgaussians}. In the color transfer step, we fix the density component of the initial 3DGS and optimize it using $\mathcal{L}{rec}$ for 200 iterations. We empirically set the filtering ratio k\% of the opacity to 5\% and that of the scale to 8\% for all scenes. We run the filter once every 100 iterations. During the stylization process, we fine-tune all parameters of $G{\theta}^{sty}$ over 800 iterations, while disabling the adaptive density control of 3DGS.
We adopt a pretrained VGG-16 network \cite{simonyan2014very} to extract the feature maps $F_{render}$, $F_{style}$, and $F_{content}$ used in the NNFM loss. The network consists of 5 convolution blocks, and we use the 'conv2' and 'conv3' blocks as the feature extractors, following the observations of the ablation study in Sec.~\ref{sec:supp_exp}.

The loss weights $(\lambda_{rec}, \lambda_{sty}, \lambda_{con}, \lambda_{sca}, \lambda_{opa}, \lambda_{tv})$ are  respectively set to $(0.2, 2, 0.005, 0.05, 0.05, 0.02)$. The depth preservation loss weight $\lambda_{dep}$ is set to $0.01$ for all forward-facing captures and $0.05$ for all 360$^\circ$ captures. At each stylization training iteration, we render an image from a sampled view different from all the training views to compute losses. The Adam optimizer \cite{kingma2014adam} is adopted with a learning rate exponentially decayed from $0.1$ to $0.01$.

\subsection{Color Transfer}
In the color transferring procedure, we employ histogram matching to adjust the color distribution of the style image to align with that of the content images. We select this transformation to ensure that the mean $\mu$ and covariance $\Sigma$ of the RGB values in the recolored training views $p_c^{re}$ match all pixels in the style image $p_s$. The transformation solution $\mathbf{A}$ and $b$ are determined to satisfy the constraints:

\begin{equation}
\begin{aligned}
b &= \mu_{p_c} - \mathbf{A} \mu_{p_s} \\
\mathbf{A}\Sigma_{p_s}\mathbf{A}^T &= \Sigma_{p_c}
\end{aligned}
\end{equation}

A family of solutions for $\mathbf{A}$ that satisfies these constraints exists. We consider utilizing the principal component analysis (PCA) method. First, let the eigenvalue decomposition of a covariance matrix be $\Sigma = \mathbf{U} \Lambda \mathbf{U}^T$. Then, we define the matrix square-root as: $\Sigma^{1/2} = \mathbf{U} \Lambda^{1/2} \mathbf{U}^T$. Subsequently, the transformation is given by:

\begin{eqnarray}
\mathbf{A} = \Sigma_{p_c}^{1/2} \Sigma_{p_s}^{-1/2}
\end{eqnarray}

The transformation solutions are also applied to the color parameter $c$ to recolor the 3DGS representation $G_{\theta}^{rec}$.

\subsection{Spatial Control}
During our spatial control process, users can acquire region masks via point-based interaction or language-based interaction. In the language-based approach, the text prompt enables the generation of masks for different perspectives using LangSAM \cite{langsam_github}. In the point-based approach, users assign several points on a single-view image, and our mask-tracking strategy facilitates the attainment of consistent multi-view masks.

Our mask-tracking strategy is based on the assumption that the area of the mask regions should be similar across different perspectives, which yields better performance in forward-facing scenes. Upon user specification of points $P_0$, SAM \cite{kirillov2023segany} is employed to obtain a mask $M_0$ for the initial view. When acquiring the mask for subsequent views, the initially specified points remain fixed. If the area of the acquired mask falls within a certain error range $\Delta s$ relative to the first mask $M_0$, it is assigned as the final mask for that view. If the discrepancy is significant, a search is conducted within the neighborhood pixels of the specified point to find a mask with an area close to the first mask. This process is repeated for all perspectives, with each subsequent view’s mask search beginning from the final specified point of the previous view. Algorithm \ref{alg:mask} illustrates the general process for our mask-tracking strategy.

\begin{algorithm}
\DontPrintSemicolon
 \caption{StylizedGS mask tracking.}\label{alg:mask}
  \SetKwFunction{SAM}{SAM}
  \SetKwFunction{Area}{Area}
  \SetKwFunction{Neighbor}{Neighbor}
  \SetKwInOut{Input}{Input}
  \SetKwInOut{Output}{Output}
  \Input{User specified points $P_0$, traing view images $\{I_i\}_{i\in [n]}$
  }
  
{$\triangleright$  \textit{Initialization}}\;
  
  $M_0 \leftarrow $ SAM$(P_0, I_0)$ 
  
{ $\triangleright$  \textit{Dynamically Tracking}}\;  
  \For{$i\leftarrow 1$ \KwTo $n-1$}{     
    $M_i \leftarrow $ SAM$(P_{i-1}, I_i)$
    
    \While{$||$Area$(M_i)-$Area$(M_0)||>\Delta s$}{
        \For{$P\leftarrow P_{i-1}$ \KwTo Neighbor$(P_{i-1})$}{
            $M_i \leftarrow $ SAM$(P, I_i)$
        }
        $P_{i-1} \leftarrow P$
     
  } }  
  \Output{$\{M_i\}_{i \in [n]}$}
\end{algorithm}

\subsection{User Study}
In the user study, we randomly selected 40 sets of stylized views of 3D scenes from both the LLFF and T\&T datasets, processed by various methods. We invited 52 participants (28 males and 24 females, aged 18 to 50). Our questionnaire was carefully designed to include participants from diverse backgrounds and age groups, reducing the risk of echo chamber effects. Participants were given ample time for unbiased evaluation, ensuring a thorough and fair assessment. Initially, we presented the participants with a style image, the original scene, and two stylized videos generated by our method and a randomly chosen comparison method. Participants were then asked to vote on the videos based on four evaluation indicators: visual preference, style match level, content preservation level, and 3D consistency quality. The specific descriptions of these four indicators are as follows:
\begin{itemize}
    \item Visual preference: Which one do you prefer in terms of visual quality?
    \item Style match level: Which one has a higher level of matching with the style image?
    \item Content preservation level: Which one maintains the content of the original scene in a better way? 
    \item 3D consistency: Which one has better 3D consistency of the scene after stylization? 3D consistency refers to whether the scene remains consistent when the camera perspective changes in the rendered video. Generally speaking, some flickering can be observed in the scene when it is inconsistent.
\end{itemize}
These descriptions are also shown to participants at the beginning of the questionnaire.

\begin{figure}
    \centering
    \captionsetup{type=figure}
    \includegraphics[width=\linewidth]{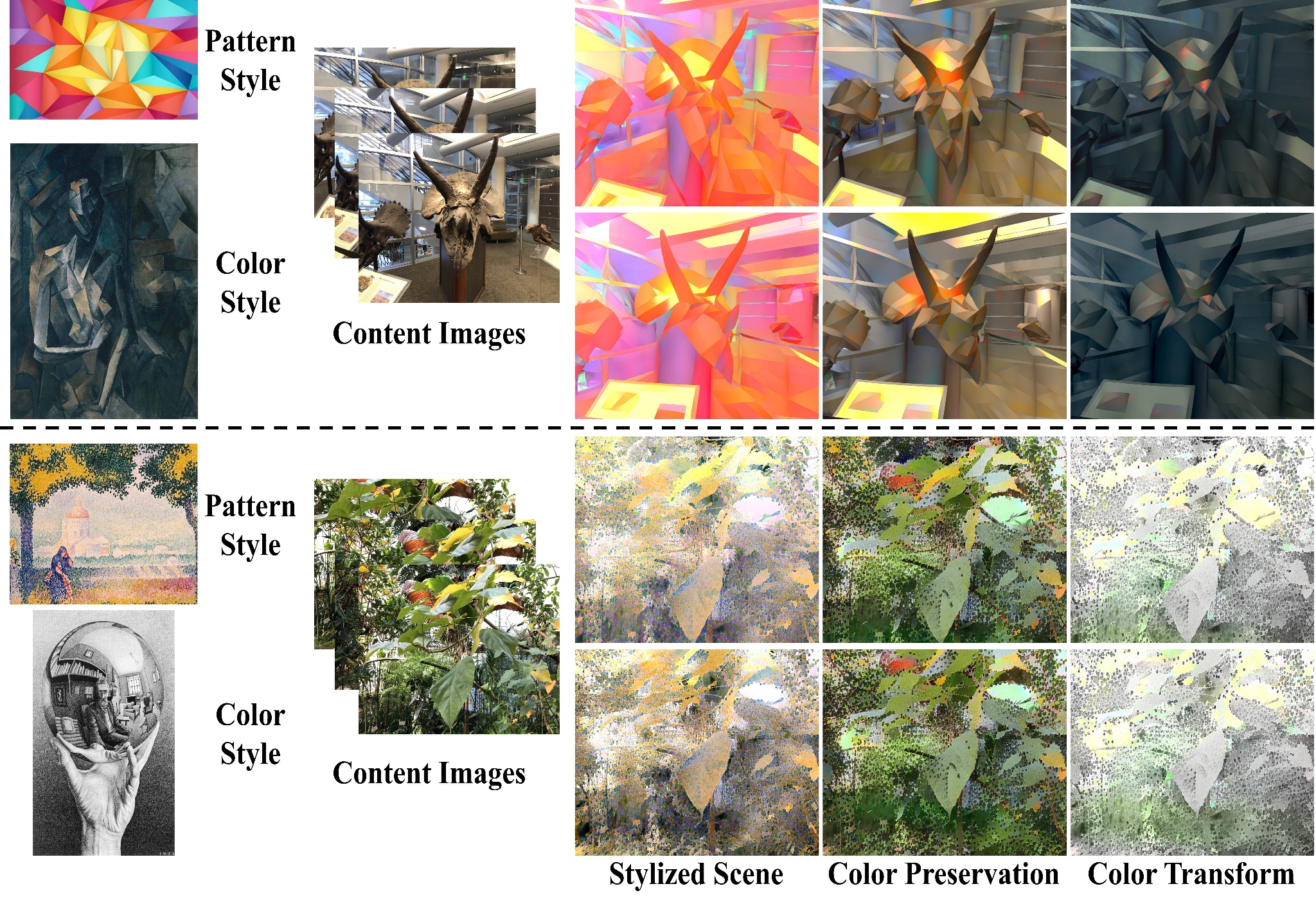}
    \captionof{figure}{\textbf{Additional color control stylization results.} Our approach facilitates versatile color management in stylized outputs, allowing users to retain the scene's original hues or apply distinct color schemes from alternative style images. Users can choose to transfer the entire style, only the pattern style, or a mix of arbitrary patterns and color styles.}
    \label{fig:addition_color}
\end{figure}

\begin{figure}
    \centering
    \captionsetup{type=figure}
    \includegraphics[width=\linewidth]{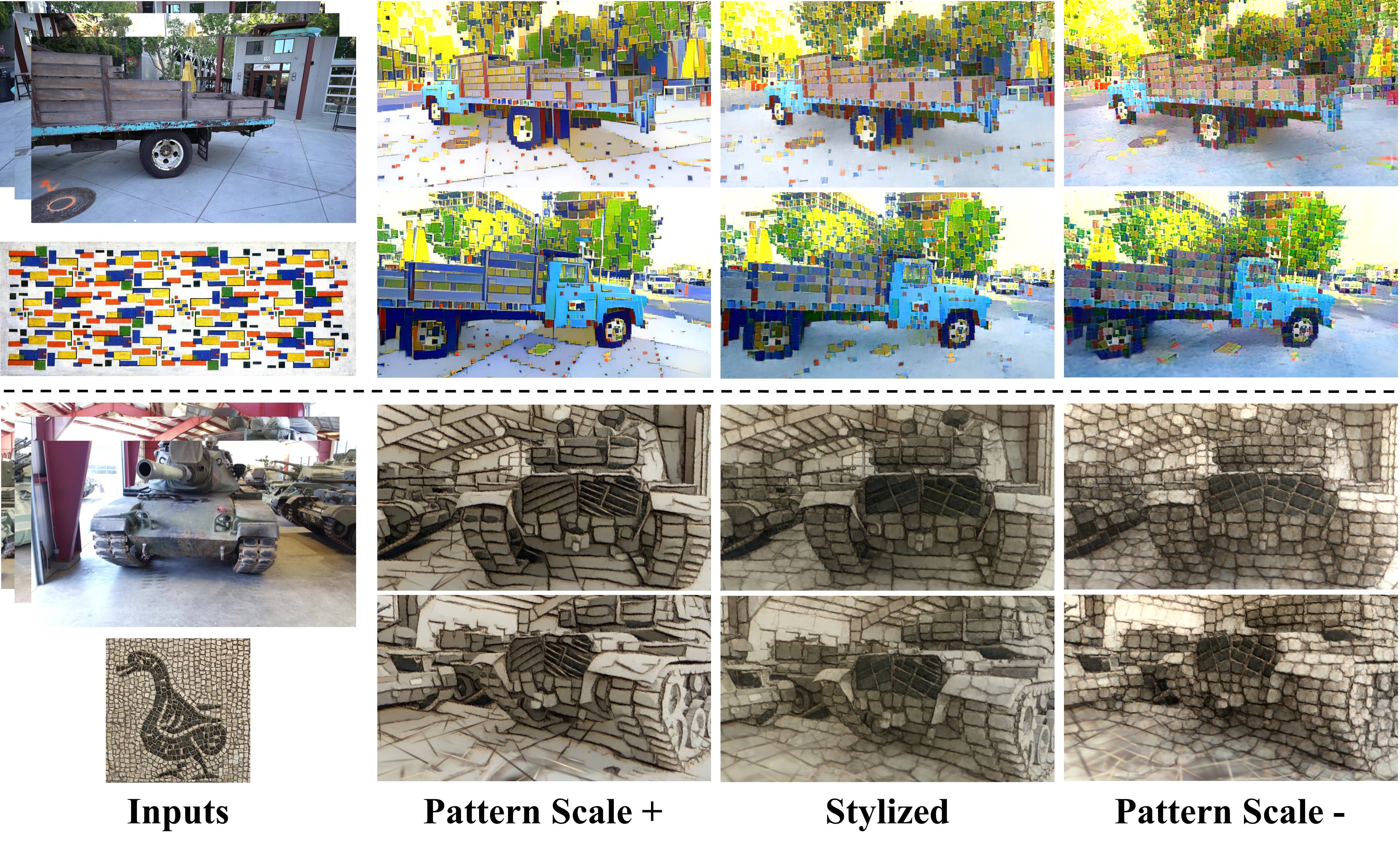}
    \captionof{figure}{\textbf{Additional scale control stylization results.} Our method enables users to flexibly control the scale of basic style elements, such as adjusting the density of colorful blocks, as demonstrated in the first row.}
    \label{fig:addition_scale}
\end{figure}

\begin{figure}
    \centering
    \captionsetup{type=figure}
    \includegraphics[width=\linewidth]{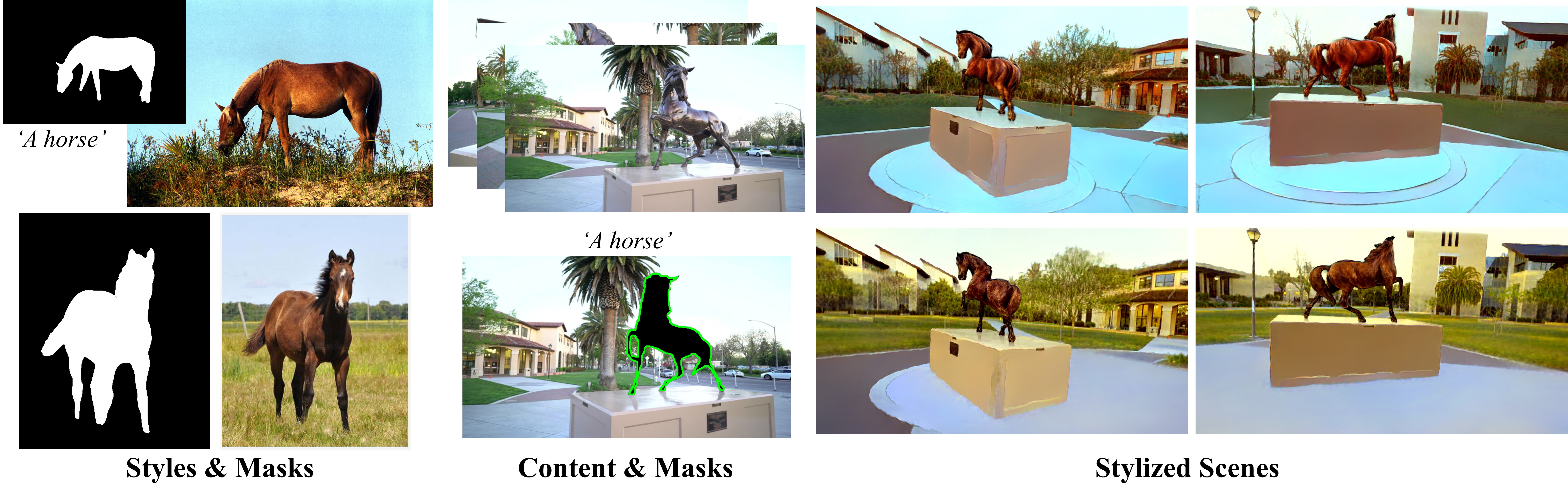}
    \captionof{figure}{\textbf{Additional spatial control stylization results.} By specifying masks with text prompt in the style image and content images, users can transfer different styles to desired regions with semantic correspondence. Note that the style images are realistic images, so the results are photo-realistic style transfer.}
    \label{fig:addition_spatial}
\end{figure}

\section{Additional Results}
\label{sec:results}
\begin{figure*}
    \centering
    \captionsetup{type=figure}
    \includegraphics[width=\linewidth]{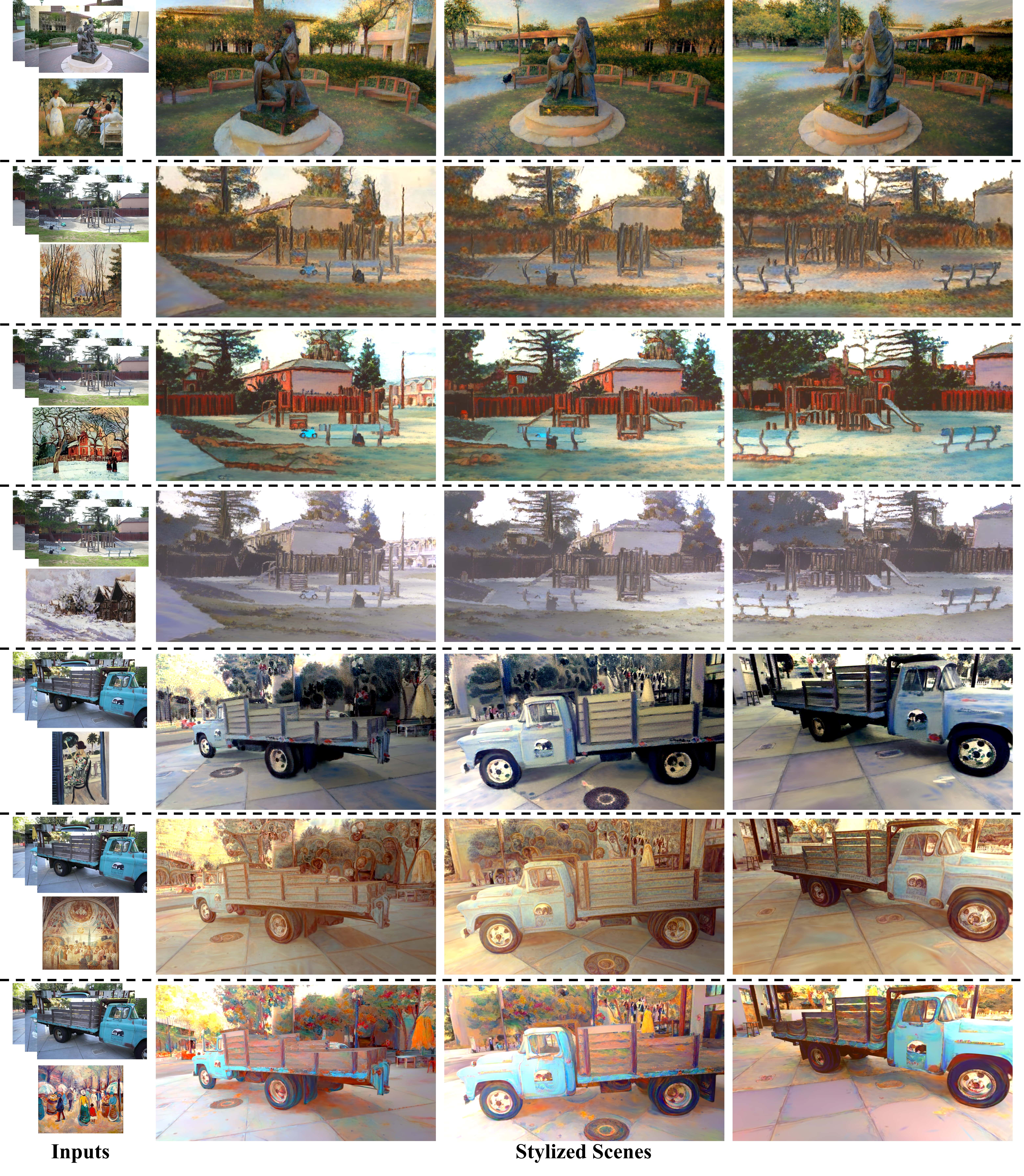}
    \vspace{-4mm}
    \captionof{figure}{\textbf{More stylization results on T\&T dataset.} It can be seen that our method faithfully captures both the color styles and pattern styles across different views.}
    \label{fig:more_tnt}
    \vspace{-4mm}
\end{figure*}

\textbf{Stylization.} 
We provide additional stylization results on the T\&T dataset in Fig.~\ref{fig:more_tnt}
These illustrations show our method's ability to faithfully capture both color and pattern styles across different views.

\textbf{Controllable Stylization.}
An advantage of our approach is the ability to perform controls over stylization. We present additional color control results in Fig.~\ref{fig:addition_color}, scale control results in Fig.~\ref{fig:addition_scale}, and spatial control results in Fig.~\ref{fig:addition_spatial}. These results demonstrate the capability of our controllable stylization method. It's noteworthy that in spatial control, the style images used are realistic, resulting in photo-realistic style transfer. Moreover, to enhance the semantic correspondence between the style image and content images, we employ LangSAM to extract masks through text input for both style and content images, facilitating the transfer of specified styles.

\section{Additional Validations}
\label{sec:supp_exp}
\begin{figure*}[ht]
    \centering
    \captionsetup{type=figure}
    \includegraphics[width=\linewidth]{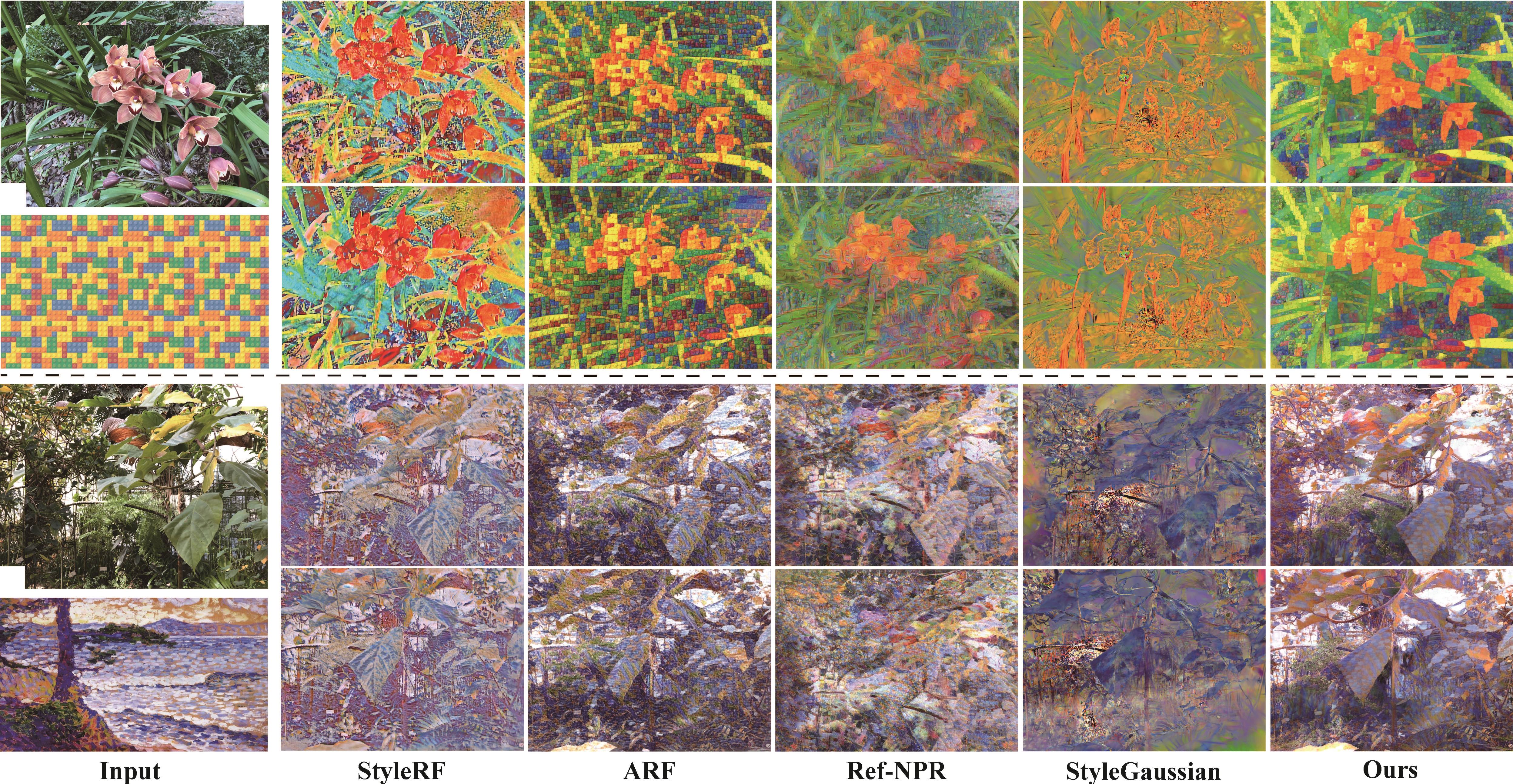}
    \vspace{-4mm}
    \captionof{figure}{\textbf{Additional comparisons with baselines on LLFF dataset.} Compared to other methods, our method excels in learning intricate and accurate geometric patterns while effectively preserving the semantic content of the original scene.}
    \label{fig:comp_llff_supp}
    \vspace{-4mm}
\end{figure*}

\begin{figure*}[ht]
    \centering
    \captionsetup{type=figure}
    \includegraphics[width=\linewidth]{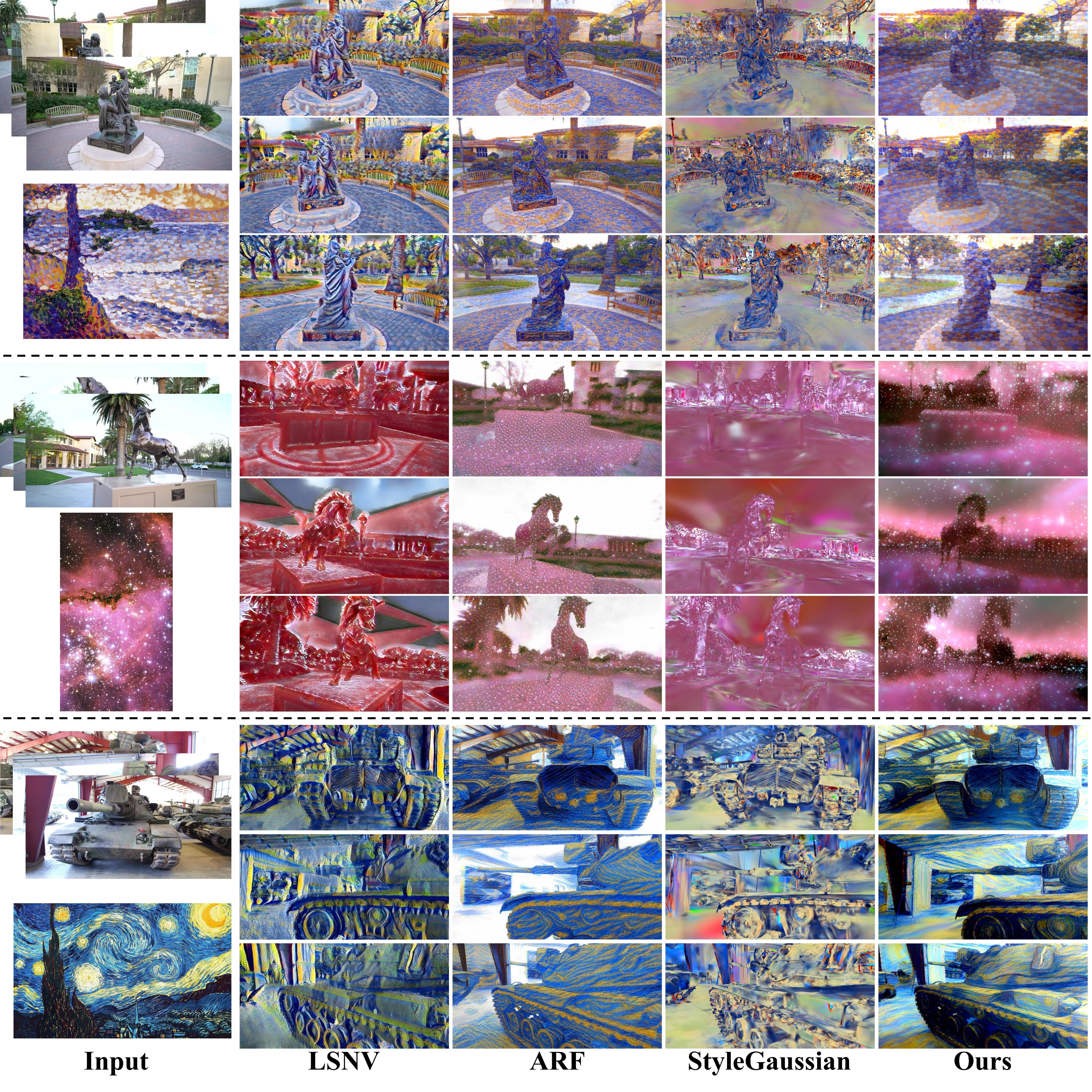}
    \vspace{-4mm}
    \captionof{figure}{\textbf{Additional comparisons with baselines on T\&T dataset.} It can be seen that our method has better results which faithfully capture both the color styles and pattern styles across different views.}
    \label{fig:comp_tnt_supp}
    \vspace{-4mm}
\end{figure*}

\textbf{More Comparisons.}
We present additional qualitative comparisons with baselines on the T\&T dataset in Fig. \ref{fig:comp_tnt_supp} and on the LLFF dataset in Fig. \ref{fig:comp_llff_supp}. It is evident from these comparisons that our method surpasses other techniques in effectively transferring both geometric patterns and color styles.

\begin{figure*}
    \centering
    \captionsetup{type=figure}
    \includegraphics[width=\linewidth]{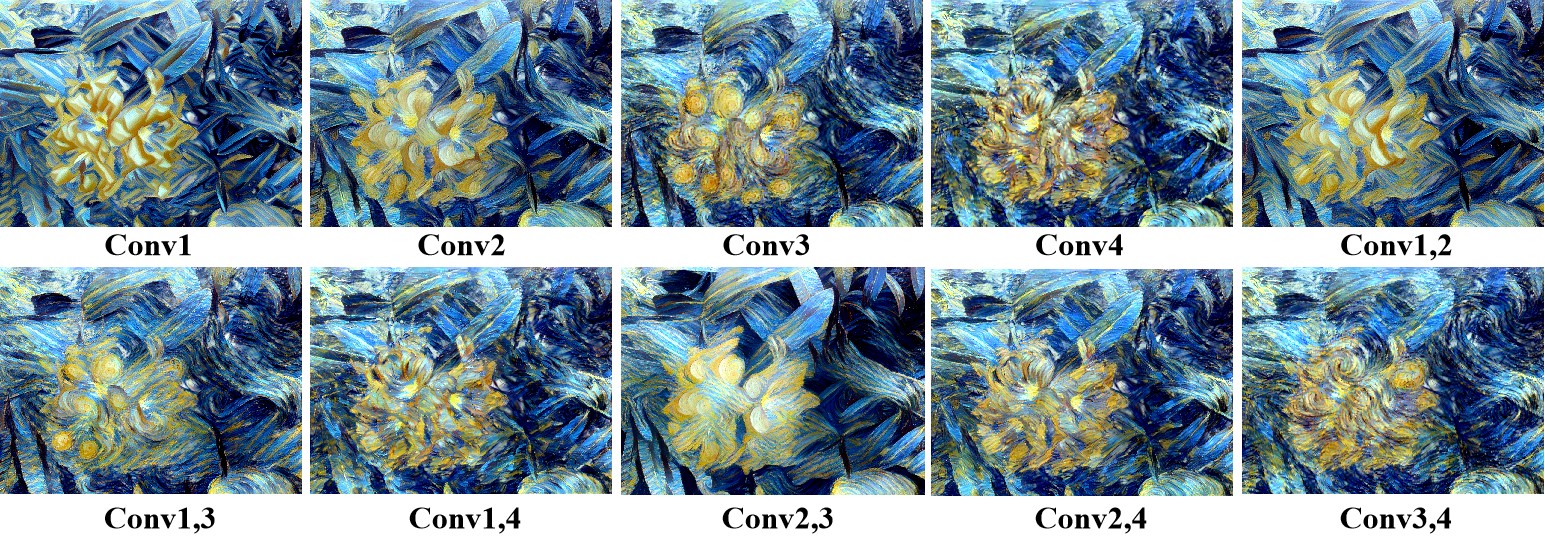}
    \vspace{-4mm}
    \captionof{figure}{\textbf{Ablation study of feature blocks.} We visualize the stylization results of extracting features from different blocks of the VGG-16 network. It can be seen that the combination of features from `conv2' and `conv3' blocks is the best choice, which has better stylization results.}
    \label{fig:ablation_conv}
    \vspace{-4mm}
\end{figure*}

\textbf{Additional Ablation Study.}
In order to select appropriate feature maps in the calculation of the NNFM loss, we conduct an ablation study on the feature maps extracted from different blocks of the VGG-16 network, as shown in Fig.~\ref{fig:ablation_conv}. From the comparison, it can be seen that selecting both `conv2’ and `conv3’ blocks for feature extraction preserves style details better than other alternatives.

\section{Future Work}
\label{sec:future}
Reconstructed 3DGS representation may include some floaters, which do not have large impact for scene rendering. However, they can significantly disrupt our stylization process. To address this issue, we propose a filter-based refinement approach. Nevertheless, it is worth noting, as mentioned in the limitations, that this process cannot entirely eliminate the impact of floaters.

In future work, we intend to explore the integration of recent advancements in Gaussian reconstruction methods~\cite{Guedon2023SuGaRSG,huang20242d}. These methods incorporate regularization terms or compress the 3D volume into a series of 2D oriented planar Gaussian disks, thereby mitigating the occurrence of floaters and improving overall reconstruction quality.
On the other hand, we will explore feed-forward-based 3DGS stylization methods, especially controllable stylization, which need to overcome the challenge between sparse user interactions and parameter changes of dense 3D Gaussians.

Finally, current language-based editing and stylization methods provide users with a new way of interaction. We will further explore how to implement language-based controllable stylization on 3DGS.

\end{document}